\documentclass[10pt,twocolumn,letterpaper]{article}

\usepackage{cvpr}              % To produce the CAMERA-READY version
\usepackage{etoolbox}
\usepackage{xparse}

\let\origsection\section
\RenewDocumentCommand{\section}{s o m}{%
  \vspace{-0.45em}%
  \IfBooleanTF{#1}
    {\origsection*{#3}}
    {\IfNoValueTF{#2}{\origsection{#3}}{\origsection[#2]{#3}}}%
  \vspace{-0.35em}%
}

\let\origsubsection\subsection
\RenewDocumentCommand{\subsection}{s o m}{%
  \vspace{-0.35em}%
  \IfBooleanTF{#1}
    {\origsubsection*{#3}}
    {\IfNoValueTF{#2}{\origsubsection{#3}}{\origsubsection[#2]{#3}}}%
  \vspace{-0.25em}%
}

\let\origsubsubsection\subsubsection
\RenewDocumentCommand{\subsubsection}{s o m}{%
  \vspace{-0.25em}%
  \IfBooleanTF{#1}
    {\origsubsubsection*{#3}}
    {\IfNoValueTF{#2}{\origsubsubsection{#3}}{\origsubsubsection[#2]{#3}}}%
  \vspace{-0.20em}%
}

\AtBeginEnvironment{figure}{\vspace{-0.30em}}
\AtEndEnvironment{figure}{\vspace{-0.45em}}
\AtBeginEnvironment{figure*}{\vspace{-0.30em}}
\AtEndEnvironment{figure*}{\vspace{-0.45em}}
\AtBeginEnvironment{table}{\vspace{-0.30em}}
\AtEndEnvironment{table}{\vspace{-0.45em}}
\AtBeginEnvironment{table*}{\vspace{-0.30em}}
\AtEndEnvironment{table*}{\vspace{-0.45em}}

\definecolor{cvprblue}{rgb}{0.21,0.49,0.74}
\usepackage[pagebackref,breaklinks,colorlinks,allcolors=cvprblue]{hyperref}
\usepackage{afterpage}
\usepackage{float}
\usepackage{listings}
\usepackage{tabularx}
\usepackage[accsupp]{axessibility}
\def\paperID{} % *** Enter the Paper ID here
\def\confName{CVPR}
\def\confYear{2026}

\title{HUMORCHAIN: Theory-Guided Multi-Stage Reasoning for Interpretable Multimodal Humor Generation}

\author{
Jiajun Zhang\thanks{Equal contribution.}\ \ \thanks{Work done during internship at Peking University.}\\
Peking University\\
Beijing, China\\
{\tt\small jzhang3439-c@my.cityu.edu.hk}
\and
Shijia Luo\footnotemark[1]\ \ \footnotemark[2]\\
Ocean University of China\\
Qingdao, China\\
{\tt\small luoshijia@stu.ouc.edu.cn}
\and
Ruikang Zhang\footnotemark[1]\\
Peking University\\
Beijing, China\\
{\tt\small 2300018416@stu.pku.edu.cn}
\and
Qi Su\thanks{Corresponding Author.}\\
Peking University\\
Beijing, China\\
{\tt\small sukia@pku.edu.cn}
}

\begin{document}

\maketitle

\begin{abstract}
Humor, as both a creative human activity and a social binding mechanism, has long posed a major challenge for AI generation. Although producing humor requires complex cognitive reasoning and social understanding, theories of humor suggest that it follows learnable patterns and structures, making it theoretically possible for generative models to acquire them implicitly. In recent years, multimodal humor has become a prevalent form of online communication, especially among Gen Z, highlighting the need for AI systems capable of integrating visual understanding with humorous language generation. However, existing data-driven approaches lack explicit modeling or theoretical grounding of humor, often producing literal descriptions that fail to capture its underlying cognitive mechanisms, resulting in the generated image descriptions that are fluent but lack genuine humor or cognitive depth. To address this limitation, we propose HUMORCHAIN (HUmor-guided Multi-step Orchestrated Reasoning Chain for Image Captioning), a theory-guided multi-stage reasoning framework. It integrates visual semantic parsing, humor- and psychology-based reasoning, and a fine-tuned discriminator for humor evaluation, forming an interpretable and controllable cognitive reasoning chain. To the best of our knowledge, this is the first work to explicitly embed cognitive structures from humor theories into multimodal humor generation, enabling a structured reasoning process from visual understanding to humor creation. Experiments on Meme-Image-No-Text, Oogiri-GO, and OxfordTVG-HIC datasets show that HUMORCHAIN outperforms state-of-the-art baselines in human humor preference, Elo/BT scores, and semantic diversity, demonstrating that theory-driven structured reasoning enables large language models to generate humor aligned with human perception.
\end{abstract}    
\section{Introduction}

\begin{figure*}[!t]
  \centering
  \includegraphics[width=\textwidth]{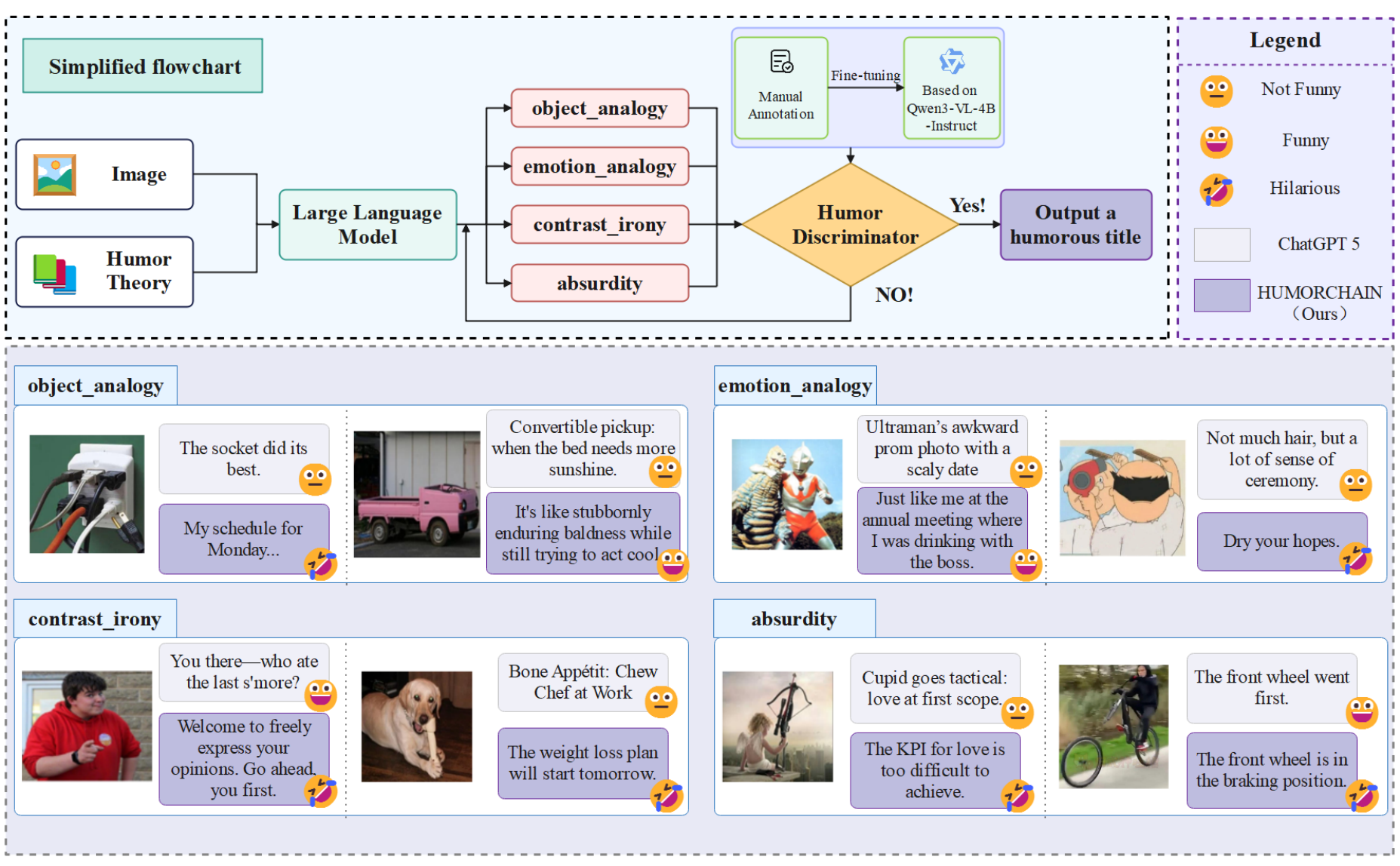} % 替换为你的图片文件名
  \vspace{-1.5em}
  \caption{ Comparison between HUMORCHAIN and ChatGPT-5 on humorous image captioning.}
  \vspace{-1em}
  \label{fig:llm_comparison}
\end{figure*}
\label{sec:intro}

%Humor plays a vital role in human communication, yet its inherent subjectivity and creativity nature make it a long-standing challenge for artificial intelligence~\cite{kim2025ai,kalloniatis2024computational}. In recent years, multimodal generative models have made remarkable progress in tasks such as image captioning~\cite{abdulgalil2025next}, producing fluent and semantically aligned text. However, when the goal shifts from accurate description to creative humor generation, these models reveal their fundamental limitations.

Humor is a fundamental component of human communication, yet its inherent subjectivity and creative nature make it a persistent challenge for artificial intelligence~\cite{kim2025ai,kalloniatis2024computational}. Although recent multimodal generative models have achieved impressive progress in tasks such as image captioning~\cite{abdulgalil2025next}, producing fluent and semantically coherent descriptions, they remain limited when the objective extends beyond factual depiction to creative humor generation. This shift reveals fundamental gaps in their reasoning capabilities and their capacity to capture the nuanced cognitive mechanisms underlying humor. Understanding these limitations requires examining what humor generation actually entails in a multimodal setting.

%Humorous image captioning requires not only linguistic fluency but also the perception of visual cues, the recognition of cognitive incongruity, and the engagement of affective responses~\cite{li2023oxfordtvg}. 
In such settings, generating humor involves not only linguistic fluency but also the perception of visual cues, the recognition of cognitive incongruity, and the engagement of affective responses~\cite{li2023oxfordtvg}.
Current models, however, lack the reasoning capacity to support these processes, leading to outputs with limited humorous effect and insufficient cognitive-affective grounding. Nonetheless, humor creation is not arbitrary: classical theoretical frameworks—such as Incongruity–Resolution Theory, Benign Violation Theory, and Superiority Theory—show that humor follows systematic and learnable structures~\cite{suls1972two,mcgraw2010benign,heyd1982place,freud1960jokes}, suggesting that generative models may be capable of implicitly internalizing these paradigms.

%To address these challenges, 
Building on this insight, we propose an innovative framework—\textbf{HUMORCHAIN} (HUmor-guided Multi-step Orchestrated Reasoning Chain for Image Captioning). By integrating established humor and psychological theories~\cite{suls1972two,mcgraw2010benign,heyd1982place,freud1960jokes,yus2021incongruity,varela2023looking}, we design a theory-guided multi-stage reasoning framework. Specifically, the framework first identifies visual cues and potential humor triggers within an image, then applies corresponding humor theories and generation strategies according to image type, thereby producing captions consistent with the cognitive mechanisms of humor perception, as shown in Figure~\ref{fig:llm_comparison}.

Given the inherently subjective nature of humor evaluation, traditional automated metrics struggle to accurately assess generation quality. To address this issue, we introduce a human-preference-guided humor discrimination mechanism. We fine-tune the Qwen3-VL-4B-Instruct model~\cite{yang2025qwen3} using instruction tuning and attach a classification head to construct a humor discriminator capable of reliably determining whether a generated caption is humorous, thereby forming a closed-loop feedback system with the generation module.

Extensive experiments on the Meme-Image-No-Text, Oogiri-GO~\cite{zhong2024let}, and OxfordTVG-HIC~\cite{li2023oxfordtvg} datasets show that HUMORCHAIN substantially outperforms existing baselines in human humor preference, Elo/BT scores, and semantic diversity. These results further confirm that integrating humor theory can effectively enhance the ability of large language models to generate humor that aligns with human perception.

Our main contributions are threefold:
\begin{enumerate}
    \item We propose \textbf{HUMORCHAIN} that systematically integrates humor and psychological theories into multimodal generation, realizing theory-driven structured humor reasoning.
    \item We design a human-preference-guided humor discriminator, forming a closed-loop ``generate--evaluate--refine'' optimization framework that significantly improves humor quality and stability.
    \item We construct a human-annotated humor preference dataset containing over 5{,}000 image--caption pairs, advancing evaluation and modeling in humor generation.
\end{enumerate}

%-------------------------------------------------------------------------

\section{Related Work}
\label{sec:formatting}
\subsection{Humorous Image Captioning and Multimodal Reasoning}

Current research on humorous image captioning mainly follows two directions: data-driven and strategy-oriented approaches. %The former relies on large-scale corpora, such as the OxfordTVG-HIC dataset~\cite{li2023oxfordtvg}, with 2.9 million image-text pairs, humor ratings, and emotional diversity annotations for content safety and diversity. MemeCraft~\cite{wang2024memecraft} also uses large-scale training for visual-humor alignment. These methods suffer from high data dependency and limited stylistic diversity. These methods suffer from high data dependency and limited stylistic diversity.
Data-driven models rely heavily on large-scale corpora, such as the OxfordTVG-HIC dataset~\cite{li2023oxfordtvg}, which contains 2.9 million image-text pairs with humor ratings and emotional annotations, and MemeCraft~\cite{wang2024memecraft}, which performs large-scale training for visual-humor alignment. However, these methods suffer from strong data dependence and tend to inherit the distributional biases of their training data, resulting in limited stylistic diversity.

Strategy-oriented research focuses on optimizing generation logic.
Early unimodal methods (e.g., MemeBot~\cite{sadasivam2020memebot}) use text template generation, overlooking humor-relevant visual cues. More recent multimodal approaches (e.g., XMeCap~\cite{chen2024xmecap}) integrate image segmentation and visual feature extraction, optimizing modality alignment and generation quality via supervised fine-tuning and reinforcement learning. However, their reliance on fixed prompt templates restricts dynamic reasoning and deep humor understanding. The CLoT model~\cite{zhong2024let} introduces ``Leap-of-Thought'' creative thinking to enhance novelty and surprise, partially improving stylistic diversity, but its outputs still fall short of producing coherent and genuinely humorous content with consistent overall quality. 
Some methods have focused on humor theories. For example, OxfordTVG-HIC~\cite{li2023oxfordtvg} employs humor theories to evaluate generated titles. However, existing work lacks exploration of their integration into the generation process, leaving their contribution to humor generation underexplored. %added

Multimodal Large Language Models (MLLMs) have made notable progress in vision-language integration. Early research established foundational architectures, training strategies, and evaluation protocols~\cite{yin2024survey,wu2023multimodal}. Subsequent models extended these foundations to support complex multimodal reasoning tasks such as image captioning, visual question answering, and cross-modal generation~\cite{song2023bridge,liang2024comprehensive,han2025multimodal}. Building on this progress, generalization has further improved through few-shot adaptation, visual prompting, and modality expansion~\cite{huang2024surveyevaluationmultimodallarge,li2025survey,guo2025mmrl,mitra2025enhancingfewshotvisionlanguageclassification}. Parallel lines of research explore interpretability, structured reasoning, and cross-modal transfer learning~\cite{caffagni2024revolutionmultimodallargelanguage,wu2024personalizedmultimodallargelanguage,cui2023surveymultimodallargelanguage,jin2024efficientmultimodallargelanguage,song2025bridge,najdenkoska2023meta}, thereby broadening the applicability of MLLMs to creative generation, situational understanding, and knowledge discovery.

Among the reasoning techniques empowered by MLLMs, Chain-of-Thought (CoT) stands out by enhancing step-by-step reasoning that improves interpretability~\cite{wei2022chain,kojima2022large,wang2022self}. In humor generation, Chen et al. proposed the text-based Chain-of-Humor (CoH), which generates humorous sentences through concept extraction and conflict insertion~\cite{chen2024xmecap}. Most existing multimodal studies, however, follow a shallow pipeline from visual description to joke generation~\cite{weller2019humor,hasan2019ur}, constrained by the lack of humor-theoretic guidance and inadequate modeling of visual cues.

\subsection{Humor Theory}
Humor is grounded in four foundational cognitive and psychological theories:

\textbf{Incongruity--Resolution Theory} is the dominant paradigm in contemporary humor studies~\cite{raskin1985semantic}. From a cognitive perspective, it defines humor as the perception and subsequent resolution of incongruity between conceptual elements~\cite{ritchie2004linguistic,palmer2003taking}. The humorous experience therefore unfolds in two stages: the individual first perceives an inconsistency, then cognitively reconstructs its internal logic to achieve resolution~\cite{suls1972two}. Building on this, Attardo and Raskin proposed the Script-based Semantic Theory of Humor (SSTH) and later the General Theory of Verbal Humor (GTVH), significantly enhancing explanatory power~\cite{attardo1991script}.

\textbf{Benign Violation Theory}, proposed by McGraw and Warren~\cite{mcgraw2010benign}, defines humor as the result of a ``benign breach'' of norms. It posits that humor arises only when two conditions are met simultaneously: (1) violation---a mild transgression of social or cognitive expectations---and (2) benignity---contextual cues mitigating the perceived harm. The dynamic balance between these dimensions determines the strength of the humorous effect.

\textbf{Superiority Theory}, originating from Hobbes~\cite{heyd1982place}, conceptualizes laughter as a psychological response to the sudden awareness of one’s superiority. Its limitations lie in its narrow explanatory range (failing to capture all humor types) and neglect of incongruity as a core feature~\cite{cai2005}. Contemporary research has expanded its scope by incorporating nonhuman targets and self-deprecating humor, broadening its applicability.

\textbf{Relief Theory}, first proposed by Freud and later elaborated by Spencer~\cite{freud1960jokes,spencer1870principles}, diverges from the others by focusing on humor’s function. It interprets humor as a psychological release mechanism---a safe outlet for repressed emotions constrained by social norms. Spencer further argued that laughter serves as the discharge of surplus nervous energy, triggered by tension between cognitive expectation and external stimuli.

In summary, these four foundational theories reveal humor’s cognitive architecture and psychological dynamics from complementary perspectives. Taken together, they provide the conceptual basis for embedding humor cognition into multimodal generative models and constructing interpretable reasoning frameworks for humor production.

\section{HUMORCHAIN Framework}

\subsection{Theory-Guided Multi-Stage Reasoning Framework}

\begin{figure}[ht]
  \centering
  \includegraphics[width=\columnwidth]{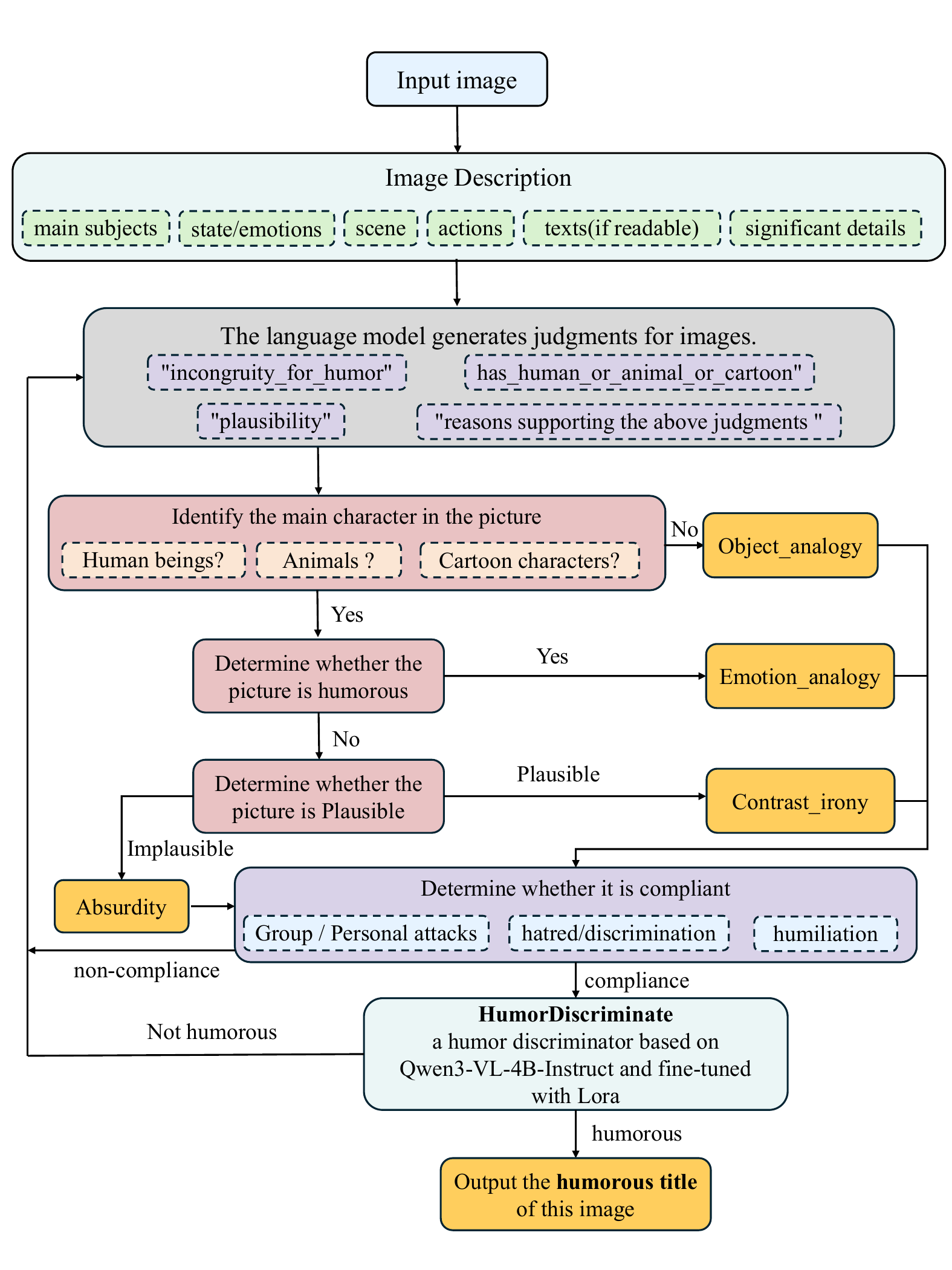} % 使用 \columnwidth 而不是 \textwidth
  \vspace{-2em}
  \caption{Workflow of the proposed \textbf{HUMORCHAIN} framework.}
  \vspace{-1.5em}
  \label{fig:Workflow}
\end{figure}

The core challenge of humorous image captioning lies in operationalizing abstract humor theories into executable structured reasoning procedures. HUMORCHAIN tackles this challenge through a multi-stage LLM reasoning framework that systematically integrates humor-theoretic and psychological mechanisms into each reasoning step.

Building upon the cognitive mechanisms outlined in classical humor theories, HUMORCHAIN formalizes humor generation as a sequential reasoning chain.  The model first detects incongruous information in the visual domain (corresponding to Incongruity--Resolution Theory~\cite{suls1972two}), then introduces irony or self-deprecation to stimulate emotional engagement (Superiority Theory~\cite{heyd1982place}), or intentionally violates norms in a controlled manner (Benign Violation Theory~\cite{mcgraw2010benign}). Finally, the model achieves affective release through linguistic expression (Relief Theory~\cite{freud1960jokes,spencer1870principles}), as shown in \autoref{fig:Workflow}. This mapping transforms humor theories into a multimodal reasoning architecture, granting the humor generation process explicit cognitive logic and theoretical interpretability.

%增加
HUMORCHAIN decomposes the humor generation process into multiple stages, each assigned a specific cognitive task to systematically progress from visual perception to humorous caption creation. Following Yus's (2021) classification of image macro memes~\cite{yus2021incongruity}, the framework adapts and generalizes the categories into four humor generation strategies:

%\begin{enumerate}
\noindent \textbf{1. Absurdity}: Rooted in evolutionary psychology, studies (e.g., Varela et al.~\cite{varela2023looking}) show that humans prioritize processing of action and emotional information in living entities. Thus, captions should primarily focus on the emotional or behavioral states of people, animals, or characters in the image. According to Incongruity--Resolution Theory, humor arises when viewers perceive an incongruity and subsequently resolve it through reinterpretation. A caption providing explanations can help individuals resolve incongruities, thereby creating humor. Therefore, for images with obvious incongruent elements, humor can be generated by producing personified descriptions based on incongruity and absurdity, or by imagining subsequent scenarios around the incongruent points (e.g., speculating ``what might happen next'').

\noindent \textbf{2. Contrast\_irony}: When visual incongruity is not explicitly present, humor can be induced through semantic contrast or irony. Based on Incongruity Theory, captions can intentionally oppose the core sentiment or behavior depicted in the image, creating an absurd reversal. Alternatively, irony can be constructed under Superiority Theory, where the combination of caption and image produces an ironic effect. Benign Violation Theory constrains this process: the semantic ``violation'' intensity must stay within a safe emotional boundary---breaking norms enough to evoke mild discomfort but avoiding genuine offense---thereby producing pleasure through simultaneous perception of ``violation'' and ``safety.''

\noindent \textbf{3. Emotion\_analogy}: For images already containing humorous or emotionally charged elements, the model performs cognitive reconstruction and emotional release based on Incongruity--Resolution and Relief Theories. It identifies emotional tension or implicit humor in the visual content, analogizes it with human psychological responses in similar situations, and generates empathetic humorous expressions. The tension relief through contrast and analogy yields both cognitive insight and emotional satisfaction.

\noindent \textbf{4. Object\_analogy}: For images dominated by inanimate objects, where direct emotional or behavioral inference is difficult, the model employs object analogy, combining Superiority and Relief Theories. It extracts salient physical or contextual features and maps them to human life events or mental states (e.g., ``a messy desk'' $\rightarrow$ ``my brain before a deadline''), evoking self-deprecating or relatable humor that simultaneously expresses stress and releases it.
%\end{enumerate}

To ensure safety, HUMORCHAIN incorporates a compliance detection module to filter group or personal attacks, hate speech, or humiliating expressions (including metaphorical references). When a violation is detected, the system triggers an automatic rewriting process.

\begin{figure*}[!t]
  \centering
  \includegraphics[width=\textwidth]{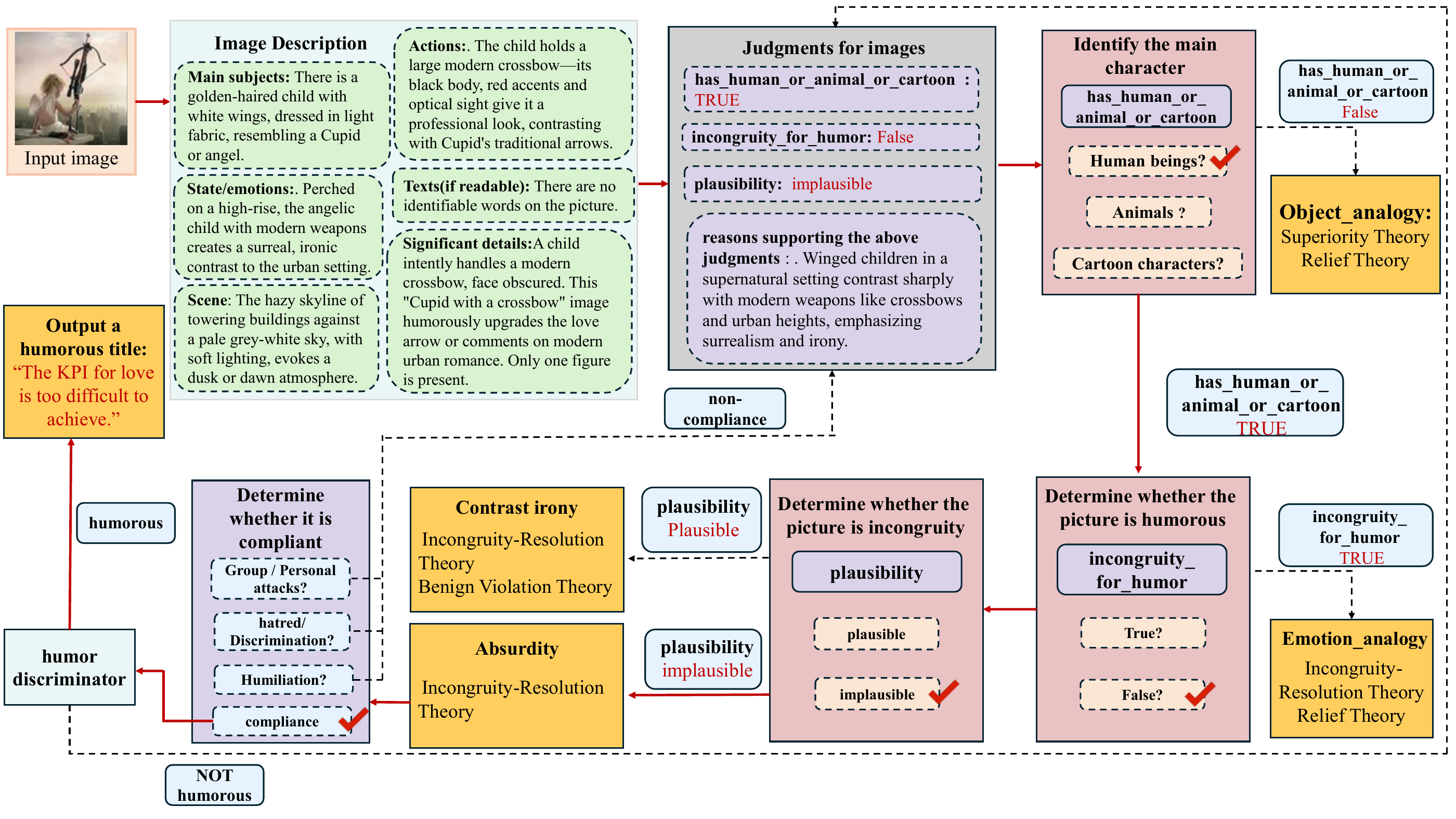} 
  \vspace{-2em}
  \caption{End-to-end example of the proposed HUMORCHAIN framework. Given an input image, HUMORCHAIN first performs visual entity recognition and determines that the scene contains a person. It then evaluates whether the image exhibits humorous characteristics, concluding that it is not humorous. Next, the system assesses the plausibility of the scene and judges it as implausible. Based on this judgment, HUMORCHAIN activates the Absurdity reasoning pathway, guided by Incongruity--Resolution Theory, to generate a corresponding humorous caption that aligns with this type of humor. The red arrow indicates the path through which the caption is generated, while the black dashed arrows represent other possibilities.}
  \vspace{-1em}
  \label{fig:wide_figure}
\end{figure*}

Finally, the framework employs a Humor Discriminator to perform binary classification (``humorous'' vs. ``non-humorous'') on generated image-caption pairs. When an output is judged as non-humorous, the system feeds back into the rewriting stage and optimizes generation strategies through controlled semantic perturbations. The discriminator’s design and training are detailed in \autoref{sec:discriminator}.
%增加
An end-to-end implementation example of the entire HUMORCHAIN reasoning pipeline is presented in Fig. \ref{fig:wide_figure}

\subsection{Humor Discriminator: Dataset and Training}
\label{sec:discriminator}

To align generated humorous captions with human preferences, we developed a lightweight humor discriminator based on Qwen3-VL-4B-Instruct~\cite{yang2025qwen3}, fine-tuned via LoRA using a small, high-quality, human-labeled dataset. The discriminator performs binary classification on image--caption pairs (``humorous'' vs. ``non-humorous''). If a caption is classified as non-humorous, the system automatically triggers a rewriting mechanism to improve alignment with desirable humor traits.

\vspace{-0.2em}
\subsubsection{Dataset Construction}
\vspace{-0.2em}

We curated a benchmark of humorous images and generated multiple candidate captions using the HUMORCHAIN framework. Each image--caption pair was independently annotated by five annotators following detailed guidelines. We defined a pair as humorous if at least two annotators labeled it as ``humorous''; otherwise, it was labeled as ``non-humorous''. This lenient threshold was chosen to retain diverse humorous expressions while filtering out clearly non-humorous cases.
%增加
This human annotation process incorporates human preference, which not only establishes a high-quality benchmark for humor discrimination but also effectively mitigates the inherent biases introduced during the operation of the prompt framework in the humor generation pipeline.

\subsubsection{Training Process}

The model was fine-tuned with LoRA using humor-aware prompts to activate implicit reasoning. Initial supervised fine-tuning enabled binary humor prediction, but this approach was inflexible. To address this, we added a classification head that outputs continuous humor probabilities, allowing adjustable thresholds for acceptance. Setting the threshold to $0.66$ provided an optimal balance between precision and regeneration cost.
%B R2/W2/W5
\subsubsection{Retry Limit and Fallback Strategy}
To ensure the efficiency and reliability of the closed-loop optimization, we set a 5-retry limit for the generation-evaluation-refine cycle, and failed cases default to the discriminator's top-rated candidate.
This enhanced architecture improves precision and stability, enabling HUMORCHAIN to generate captions better aligned with human perception.

\section{Experiments}
% added
Through systematic experiments, we evaluate HUMORCHAIN from multiple dimensions: we assess its performance in humorous image captioning; examine whether integrating humor theory with few-shot learning or knowledge-transfer reasoning enhances humor generation; and investigate the contributions of the fine-tuned humor discriminator to the overall pipeline. Specifically, we use GPT-5‑2025‑08‑07 as the backbone reasoning model in our experiments.

\subsection{Experiment 1: Pipeline Comparison and Dataset Design}

\paragraph{Data Sources}

\vspace{-1em}
\begin{table}[h]
    \centering
    %\footnotesize % 调整为较小字体（若仍偏宽可改为 \scriptsize）
    \scriptsize
    \caption{Pairwise comparisons for humorous image captioning.}
    \vspace{-1em}
    \begin{tabularx}{\columnwidth}{@{}p{1.05cm} >{\raggedright\arraybackslash}X@{}} % 自适应单栏宽度，防止越界
        \toprule
        \textbf{Pair ID} & \textbf{Comparison Objective} \\
        \midrule
        A vs B & Evaluate the benefit of few-shot prompting for humor captioning. \\
        A vs C & Assess the impact of introducing humor theory. \\
        B vs D & Explore the effect of combining theory with few-shot prompting. \\
        C vs E & Analyze the gain from CoT + theory-guided humorous image captioning. \\
        D vs F & Analyze the gain from combining CoT + theory-guided approach with few-shot prompting. \\
        E vs F & Compare the contribution of few-shot prompting within theory-guided + CoT framework. \\
        I vs A & Compare the overall improvement of HUMORCHAIN over the baseline. \\
        I vs F & Analyze the differences between HUMORCHAIN and CoT-guided methods. \\
        I vs G & Analyze the performance differences between HUMORCHAIN and CLoT. \\
        I vs H & Evaluate the performance of HUMORCHAIN under OxfordTVG-HIC dataset conditions. \\
        J vs A & Quantify the performance gain of theory-guided reasoning (without discriminator) over zero-shot baseline. \\
        J vs F & Verify the superiority of pure theory-guided multi-stage reasoning over CoT-fused baselines (without discriminator). \\
        J vs G & Compare the theory-guided framework (without discriminator) with the external CLoT model for humor generation. \\
        J vs H & Evaluate the performance of HUMORCHAIN under OxfordTVG-HIC dataset conditions. \\
        I vs J & Highlight the performance improvement brought by the discriminator and retry loop in HUMORCHAIN. \\
        \bottomrule
    \end{tabularx}
    \vspace{-0.5em}
    \label{tab:pairwise_comparisons}
\end{table}
%\vspace{-2em}

\begin{table*}[h!]
    \centering
    %\footnotesize % 调整为较小字体
    \scriptsize % 缩小字体
    \setlength{\tabcolsep}{5pt} % 缩小列间距    
    \setlength{\abovecaptionskip}{2pt}
    \renewcommand{\arraystretch}{1.05} % 略微压缩行距
    \caption{Experimental method configurations (A--I) for humorous image captioning.}
    \begin{tabularx}{\textwidth}{l l X}% 扩宽第三列，适应双栏总宽度
        \toprule
        \textbf{Group} & \textbf{Strategy} & \textbf{Description} \\
        \midrule
        A & Zero-shot & Direct image captioning without examples or theoretical cues. \\
        B & Few-shot & Incorporates example-based prompting with humor-style mimicry. \\
        C & Rule-Based & References the four theories of humor (e.g., incongruity, violation). \\
        D & Few-shot + Rule-Based & Combines examples with theoretical references to achieve structured prompts. \\
        E & Rule-Guided + CoT & Adds Chain-of-Thought reasoning to theory-guided captioning. \\
        F & Few-shot + Rule-Guided + CoT & Combines all strategies without explicit multi-stage orchestration. \\
        G & External CLoT (SYSU) & Sun Yat-sen University’s CLoT model~\cite{zhong2024let}. \\
        H & External OxfordTVG-HIC & Oxford University dataset based on humor-labeled caption pairs~\cite{li2023oxfordtvg}. \\
        I (Ours) & Theory-Guided Multi-Stage Reasoning & Proposed HUMORCHAIN framework integrating cognitive humor theory, staged reasoning, and humor discrimination feedback. \\
        %增加
        J & Theory-Guided Multi-Stage Reasoning (without discriminator) & Proposed HUMORCHAIN framework with the humor discriminator and retry loop ablated, only retaining cognitive humor theory and staged reasoning. \\      
        \bottomrule
    \end{tabularx}
    \vspace{-1em}
    \label{tab:exp_methods}
\end{table*}

This study utilizes three categories of datasets to facilitate both internal and external comparative analyses. For all internal methods (A--F), the Meme-Image-No-Text dataset is employed to ensure a consistent evaluation setting for images. For cross-model comparisons, two external evaluation sets are constructed: Oogiri-GO (Group G)~\cite{zhong2024let} and OxfordTVG-HIC (Group H)~\cite{li2023oxfordtvg}, with captions generated in parallel by HUMORCHAIN and the respective external systems. Detailed dataset configurations are provided in Appendix \ref{appx:datasource}.

A total of nine generation methods (A--I) were designed to systematically test different strategies for humor captioning, as shown in Table~\ref{tab:exp_methods}. Groups A--F are internal baselines developed in this study, while G (CLoT) and H (OxfordTVG-HIC) serve as external comparison models. I (Ours) corresponds to HUMORCHAIN, integrating theory-guided multi-stage reasoning and discriminator-based feedback.
\vspace{-1em}
\paragraph{Experimental Design and Evaluation Metrics}
Given the subjectivity of humorous image captioning, we adopt a complementary two-stage evaluation strategy, comprising Pairwise Comparison and Single-Title Evaluation. Detailed annotator instructions are provided in the Appendix \ref{sec:human_annotation_guide}.
\vspace{-1em}
\paragraph{(1) Pairwise Comparison}
To systematically assess the relative performance of diverse generation strategies, ten comparison groups were constructed to cover all key variable combinations, as summarized in Table~\ref{tab:pairwise_comparisons}.
Four metrics were derived from annotator judgments. Win Rate and Hard Win Rate quantify the proportion of preferred captions, with Hard Win Rate excluding ties. Significance tests assess the reliability of these differences. Additionally, the Bradley–Terry and Arena Elo~\cite{elo1978rating} models are used to derive relative rankings.
\vspace{-1em}
\paragraph{(2) Single-Caption Evaluation}
Annotators independently rate each caption, using a binary label to assess its humor performance. In addition, we compute a series of automated metrics: Distinct-1/2 for lexical diversity~\cite{li2016diversity}, BERTScore for semantic similarity~\cite{zhang2020bertscore}, CLIPScore~\cite{radford2021learning} for visual–semantic alignment and incongruity, and Embedding Average (EA) and Greedy Matching (GM)~\cite{liu2016not} for sentence- and word-level semantic relevance. Together, these complementary metrics form a comprehensive evaluation framework for assessing models’ humor generation capabilities.

\subsection{Experiment 2: Evaluation of the Humor Discriminator}

\paragraph{Comparison Before and After Fine-Tuning}
\label{sec:comparison sft}
We evaluated three discriminator variants on the validation set using identical, officially recommended generation parameters: 1) Baseline (un-tuned), 2) LoRA (binary classifier), 3) LoRA + Classifier Head (threshold = 0.66). This experiment demonstrates how humor dataset fine-tuning improves alignment with human preferences and shows the added flexibility from a trained classification head. We analyze confusion matrices and key metrics for each model.
\vspace{-2.5em}
\paragraph{Impact on the Overall Pipeline}
To assess the discriminator’s effect on pipeline performance, we compare the ratio of captions labeled as humorous in the fine-tuned dataset with the precision of LoRA + Classifier Head after integration. This directly reflects the discriminator’s impact on output quality. Confusion matrix metrics are also used to estimate the retry mechanism’s effect on inference cost.
\vspace{-1.5em}
\paragraph{Comparison with Other Large Models}
To further validate the fine-tuned discriminator’s advantages, we evaluate several larger closed-source LLMs on humor detection, including Gemini-2.5-Flash, GPT-4-1, and Claude-3-5-Haiku-20241022, using recommended parameters. As these models do not support training a classification head, we use a unified 0/1 binary classification with structured outputs. Metrics from Section~\ref{sec:comparison sft} are used to ensure consistency in comparison.

\section{Results and Discussion}

\subsection{Pairwise Comparison Analysis}

As shown in Table~\ref{tab:pairwise_results} and Table~\ref{tab:global_ranking}, across all nine experimental configurations (A--J), the proposed HUMORCHAIN (Method I) achieves the best overall performance. In cross-model evaluations, HUMORCHAIN consistently outperforms CLoT (Group G)~\cite{zhong2024let} and OxfordTVG-HIC (Group H)~\cite{li2023oxfordtvg}. 
%增加
Notably, the ablation model J (without discriminator) also outperforms most baselines, validating the core contribution of humor theory guidance.
These results indicate HUMORCHAIN’s robustness in humorous caption generation.

\begin{table}[h]
    \centering
    \scriptsize
    \setlength{\tabcolsep}{4pt}
    \setlength{\abovecaptionskip}{2pt}
    \renewcommand{\arraystretch}{1.05}
    \vspace{-0.5em}
    \caption{Pairwise win rate and significance analysis for humorous image captioning methods.}
    \begin{tabular*}{\columnwidth}{@{\extracolsep{\fill}} l c c c}
        \toprule
        \textbf{Comparison} & \textbf{Total} & \textbf{Win Rate A} & \textbf{Win Rate B} \\
        \midrule
        A vs B & 300 & 0.495 & 0.505 \\
        A vs C & 300 & 0.502 & 0.498 \\
        B vs D & 300 & 0.492 & 0.508 \\
        C vs E & 300 & 0.465 & 0.535 \\
        D vs F & 300 & 0.523 & 0.477 \\
        E vs F & 300 & 0.482 & 0.518 \\

        J vs A & 300 & 0.850 & 0.150 \\
        J vs F & 300 & 0.830 & 0.170 \\
        J vs G & 300 & 0.626 & 0.374 \\
        J vs H & 300 & 0.788 & 0.212 \\

        I vs A & 300 & 0.695 & 0.305 \\
        I vs F & 300 & 0.680 & 0.320 \\
        I vs G & 794 & 0.683 & 0.317 \\
        I vs H & 1007 & 0.860 & 0.140 \\

        I vs J & 300 & 0.745 & 0.255 \\
        %这里还没加上胜率
        
        \bottomrule
    \end{tabular*}
    \vspace{-1em}
    \label{tab:pairwise_results}
\end{table}

In the comparison between A and I, the results show that explicit multi-stage reasoning and humor-mechanism modeling provide a substantial advantage in humor generation. The comparison between F and I further demonstrates that HUMORCHAIN markedly outperforms CoT-based approaches in both stability and humor coherence, underscoring that effective humor generation benefits not only from explicit knowledge but also from cognitive modeling embedded in the reasoning process.

In cross-model and cross-dataset evaluations, HUMORCHAIN achieves consistently higher humor win rates in both I-vs-G and I-vs-H, indicating that Method I (Ours) maintains robust and stable humor performance across diverse conditions.

Regarding the Arena Elo (Elo) and Bradley--Terry (BT) scores (Table~\ref{tab:global_ranking}), Method I achieves the highest values on both metrics. The results further reveal a general trend in which model performance improves with increasing prompt complexity and reasoning depth. The A--C groups indicate that single-strategy enhancements yield limited gains, while implicit chain-of-thought methods (E, F) provide moderate improvements yet remain inferior to I (Ours). External models G (CLoT) and H (OxfordTVG-HIC) obtain comparatively lower scores. Overall, the global statistics reinforce HUMORCHAIN’s robustness and consistent superiority across diverse generation strategies.

\begin{table}[h]
    \centering
    \scriptsize % 字体缩小，比 \footnotesize 更小一级
    \setlength{\tabcolsep}{4pt} % 缩小列间距
    \setlength{\abovecaptionskip}{2pt}
    \renewcommand{\arraystretch}{1.05} % 略微压缩行距
    \vspace{-0.5em}
    \caption{Global ranking metrics for humorous image captioning frameworks. Arena Elo and Bradley--Terry (BT) scores are calculated from all human comparison results.}
    \begin{tabular*}{\columnwidth}{@{\extracolsep{\fill}} l c c}
        \toprule
        \textbf{Framework} & \textbf{Elo} & \textbf{BT} \\
        \midrule
        I (Ours) & 1554.60 & 3.57 \\
        F (Few-shot + CoT + Rule-Guided) & 1528.84 & 1.00 \\
        E (CoT + Rule-Guided) & 1526.46 & 0.88 \\
        D (Few-shot + Rule-Based) & 1505.43 & 0.73 \\
        C (Rule-Based) & 1504.45 & 0.54 \\
        A (Zero-shot) & 1498.09 & 0.52 \\
        B (Few-shot) & 1495.05 & 0.46 \\
        H (OxfordTVG-HIC) & 1467.40 & 0.55 \\
        G (CLoT) & 1464.06 & 0.74 \\
        \bottomrule
    \end{tabular*}
    \vspace{-1em}
    \label{tab:global_ranking}
\end{table}

\subsection{Single-Title Evaluation}

\begin{table*}[h]
    \centering
    \scriptsize % 比 \footnotesize 更小，节省空间
    \setlength{\tabcolsep}{3pt} % 缩小列间距
    \setlength{\abovecaptionskip}{2pt}
    \renewcommand{\arraystretch}{1.05} % 轻微压缩行距
    \caption{Single-Title Evaluation results for humorous image captioning. Metrics include human humor score, CLIPScore, Embedding Average (EA-Rev), Greedy Matching (GM-Rev), Distinct-1/2, and BERT Cross Score.}
    \begin{tabular*}{\textwidth}{@{\extracolsep{\fill}} l c c c c c c c}
        \toprule
        \textbf{Framework} & \textbf{Humor Mean} & \textbf{CLIPScore} & \textbf{EA-Rev} & \textbf{GM-Rev} & \textbf{Distinct-1} & \textbf{Distinct-2} & \textbf{BERT Cross Score} \\
        \midrule
        A (Zero-shot) & 0.412 & 0.630 & 0.567 & 0.458 & 0.804 & 0.988 & 0.837 \\
        B (Few-shot) & 0.418 & 0.630 & 0.560 & 0.458 & 0.757 & 0.966 & 0.833 \\
        C (Rule-Based) & 0.362 & 0.630 & 0.578 & 0.463 & 0.795 & 0.980 & 0.837 \\
        D (Few-shot + Rule-Based) & 0.394 & 0.626 & 0.609 & 0.470 & 0.728 & 0.965 & 0.836 \\
        E (CoT + Rule-Guided) & 0.382 & 0.629 & 0.557 & 0.457 & 0.750 & 0.965 & 0.838 \\
        F (Few-shot + CoT + Rule-Guided) & 0.398 & 0.625 & 0.622 & 0.466 & 0.720 & 0.937 & 0.837 \\
        G (CLoT) & 0.195 & 0.636 & 0.731 & 0.546 & 0.377 & 0.725 & 0.836 \\
        I (Ours) & 0.810 & 0.618 & 0.765 & 0.521 & 0.808 & 0.965 & 0.832 \\
        \bottomrule
    \end{tabular*}
    \vspace{-1em}
    \label{tab:single_title_eval}
\end{table*}

Combining human humor ratings with multidimensional automated metrics, a comprehensive assessment of humor quality, lexical diversity, and semantic distinctiveness was conducted (Table~\ref{tab:single_title_eval}). I (Ours) ($0.810$) significantly outperforms all baselines, while combined strategies (D, E, F) show similar performance within the $0.38$--$0.40$ range, indicating limited gains from integrating implicit reasoning and few-shot prompting. The external model G ($0.195$) performs notably worse, reflecting restricted humor generation.

Across automated metrics, I (Ours) achieves the best CLIPScore, EA-Rev, and BERT Cross Score, demonstrating stronger contrast, unexpectedness, and semantic creativity in image--text relations, consistent with incongruity-based humor mechanisms. Although G shows higher GM-Rev, suggesting more novel word usage, the overall trends are consistent: HUMORCHAIN enhances incongruity and creativity while maintaining visual--semantic relevance, achieving the strongest performance along the ``unexpectedness--humor'' dimension.

\subsection{Evaluation of the Humor Discriminator}

\subsubsection*{Comparison Before and After Fine-Tuning}

Fine-tuning with LoRA and adding a classification head both lead to substantial improvements in key metrics such as accuracy and precision (see Table~\ref{tab:discriminator_finetune}). These enhancements demonstrate that the model’s humor detection capability is significantly strengthened through targeted training and threshold adjustment.

% --- 新表格：仅保留 Precision ---
\vspace{-0.5em}
\begin{table}[h]
    \centering
    \scriptsize 
    \setlength{\abovecaptionskip}{2pt}
    \setlength{\tabcolsep}{10pt} % 适当增加列间距，因为列数变少了
    \renewcommand{\arraystretch}{1.1} 
    \caption{Humor discriminator precision performance comparison.}
    \begin{tabular*}{\columnwidth}{@{\extracolsep{\fill}} l c}
        \toprule
        \textbf{Model} & \textbf{Precision} \\
        \midrule
        Baseline & 0.523 \\
        LoRA & 0.636 \\
        LoRA + Classifier (thr = 0.66) & \textbf{0.670} \\
        \bottomrule
    \end{tabular*}
    \label{tab:discriminator_finetune}
\end{table}
\vspace{-0.75em}

\subsubsection*{Impact on Pipeline Performance}

As shown in Table~\ref{tab:discriminator_pipeline}, introducing the discriminator significantly increases the proportion of humorous outputs and overall caption quality, while maintaining a reasonable inference cost. The improved precision means users are much more likely to receive genuinely humorous outputs, representing a substantial gain in user experience. The retry mechanism ensures users receive high-quality captions, and parallel generation can further optimize efficiency.

\begin{table}[h]
    \centering
    \scriptsize
    \setlength{\tabcolsep}{3pt}
    \setlength{\abovecaptionskip}{2pt}
    \renewcommand{\arraystretch}{1.05}
    \vspace{-0.5em}
    \caption{Impact of the humor discriminator on pipeline output quality and inference cost.}
    \begin{tabular}{l c c}
        \toprule
        \textbf{Metric} & 
        \shortstack{\textbf{Before}\\\textbf{Discriminator}} & 
        \shortstack{\textbf{After}\\\textbf{Discriminator}\\\textbf{(LoRA + Classifier)}} \\
        \midrule
        Proportion of Humorous Outputs & 45.1\% & \textbf{67.0\%} \\
        Acceptance Rate (Positive Pred.) & -- & 36.5\% \\
        Avg. Generations per Accepted Caption & -- & 2.74$\times$ \\
        Precision Improvement $\Delta$ & -- & \textbf{+22\%} \\
        \bottomrule
    \end{tabular}
    \vspace{-0.75em}
    \label{tab:discriminator_pipeline}
\end{table}

\subsubsection*{Comparison with Other LLMs}

As shown in Table~\ref{tab:llm_comparison}, closed-source LLMs tend to overestimate humor, resulting in low precision. This means that simply integrating a larger or closed-source LLM does not effectively improve output quality. In contrast, our fine-tuned discriminator provides substantial gains in both accuracy and efficiency, highlighting its essential role in the pipeline.

% --- 原表格已注释 ---
% \begin{table}[H]
%     \centering
%     \scriptsize % 字体更小，比 \footnotesize 再小一级
%     \setlength{\abovecaptionskip}{2pt}
%     \setlength{\tabcolsep}{4pt} % 缩小列间距
%     \renewcommand{\arraystretch}{1.05} % 轻微压缩行距
%     \caption{Humor discriminator performance before and after fine-tuning.}
%     \begin{tabular}{l c c c c}
%         \toprule
%         \textbf{Model} & \textbf{Accuracy} & \textbf{Precision} & \textbf{Recall} & \textbf{F1} \\
%         \midrule
%         Baseline & 0.579 & 0.523 & \textbf{0.750} & 0.616 \\
%         LoRA & 0.673 & 0.636 & 0.642 & \textbf{0.639} \\
%         LoRA + Classifier (thr = 0.66) & \textbf{0.673} & \textbf{0.670} & 0.542 & 0.599 \\
%         \bottomrule
%     \end{tabular}
%     \label{tab:discriminator_finetune_old}
% \end{table}

\begin{table}[h]
    \centering
    \scriptsize % 缩小字体
    \setlength{\abovecaptionskip}{2pt}
    \setlength{\tabcolsep}{3pt} % 缩小列间距
    \renewcommand{\arraystretch}{1.05} % 压缩行距
    \vspace{-0.5em}
    \caption{Comparison of humor detection performance across LLMs.}
    \begin{tabular}{l c c c c c c}
        \toprule
        \textbf{Model} & 
        \textbf{TP} & 
        \textbf{TN} & 
        \textbf{FP} & 
        \textbf{FN} & 
        \shortstack{\textbf{Positive}\\\textbf{Rate (\%)}} & 
        \textbf{Precision} \\
        \midrule
        Gemini-2.5-Flash & 119 & 9 & 137 & 1 & 96.2 & 0.465 \\
        GPT-4-1 & 111 & 23 & 123 & 9 & 87.9 & 0.474 \\
        Claude-3.5-Haiku-20241022 & 116 & 10 & 136 & 4 & 94.7 & 0.460 \\
        \shortstack[l]{Qwen3-VL-4B-Instruct\\ (LoRA + Cls)} & 65 & 114 & 32 & 55 & \textbf{36.5} & \textbf{0.670} \\
        \bottomrule
    \end{tabular}
    \vspace{-0.75em}
    \label{tab:llm_comparison}
\end{table}

\subsection{Summary and Discussion}

Our experiments show that relying solely on zero-shot prompting, few-shot prompting, theory-guided prompts, or chain-of-thought reasoning does not substantially improve humorous caption generation. HUMORCHAIN’s explicit multi-stage reasoning pipeline integrates core mechanisms from humor studies and psychology—such as incongruity, creativity, and cognitive consistency—into the generation process, enabling the model to produce more diverse and unexpected humorous content than traditional approaches.

In addition, the fine-tuned discriminator plays a crucial role in filtering and optimizing outputs, significantly increasing the proportion of captions perceived as humorous (from $45\%$ to $67\%$). This targeted discrimination step ensures that generated captions better align with human humor perception and expectations.

These results confirm that integrating structured reasoning and targeted discrimination is essential for generating high-quality humorous content.
\section{Conclusion}

Grounded in humor and psychological theories, HUMORCHAIN integrates multi-stage reasoning with a targeted discriminator to provide a structured and interpretable framework for humor generation. The reasoning pipeline explicitly translates core mechanisms from humor and psychological theories into actionable steps, driving creative and diverse output, while the discriminator ensures quality and alignment with human expectations. 

Experiments demonstrate that HUMORCHAIN achieves consistent improvements over mainstream methods, particularly in interpretability and reasoning transparency. Its parameterized theoretical components and lightweight adaptation module also show promising potential for cross-lingual and cross-cultural transfer, enabling personalized, culture-aware, and context-aware generation. Beyond humor, the theory-driven, structured reasoning paradigm offers a universal approach for creative language tasks, such as satire and metaphor, and a general direction for multimodal generation with less data and greater control. HUMORCHAIN’s contribution lies in its theory-driven approach to structured reasoning and generation. While our framework demonstrates strong performance and improved interpretability, future work will focus on expanding its applicability to broader creative language tasks and addressing challenges in subjective evaluation and cultural adaptation.

{
    \small
    \bibliographystyle{ieeenat_fullname}
    \bibliography{main}
}

\clearpage
\setcounter{page}{1}
\maketitlesupplementary
\appendix

\section{Limitations}
Although \textsc{HUMORCHAIN} is grounded in four foundational humor theories—Incongruity–Resolution, Benign Violation, Superiority, and Relief—its theoretical coverage remains incomplete. More recent frameworks and multidisciplinary perspectives have not yet been systematically integrated, and certain forms of humor (e.g., culturally embedded irony or reference-dependent implicit jokes) may therefore be underrepresented or insufficiently captured. However, the modular design of our Theory-Guided Multi-Stage Reasoning Framework enables us to incorporate new generation strategies with only modification of generation prompts, which highlights the flexibility of our framework. In future works, we plan to incorporate and compare additional humor theories to further strengthen the theoretical grounding of the reasoning module and to more comprehensively cover diverse humor phenomena.
%增加
Humor theory-guided reasoning is inherently unable to cover all humor phenomena exhaustively, which also reflects the inherent limitation of theory-guided Chain-of-Thought reasoning; future practical deployment will therefore focus on balancing the comprehensiveness of humor theory integration and the computational complexity of the reasoning system.

Moreover, humor is inherently subjective: even for the same joke or image caption, differences in cultural background, linguistic experience, and personal preference can lead to markedly different perceptions of funniness. Although we employ pairwise comparisons and binary annotations to approximate consensus-style labels and train the discriminator accordingly, the current framework remains biased toward captions that appeal to the majority rather than modeling individual humor preferences in a fine-grained manner. Future work will explore the integration of personalization signals and user modeling to better capture such subjective variability.
%增加
\section{Validation of Selected Humor Theories}
\noindent
We validate the four adopted humor theories (Incongruity–Resolution, Benign Violation, Superiority, Relief) via quantitative analysis of 100 OxfordTVG-HIC samples, finding they apply to 86\% of cases (72\% with a single dominant theory), verifying their universality for multimodal humor.
Statistical analysis of 25\% of our generated captions shows the four corresponding generation strategies distribute as: Absurdity (26.9\%), Contrast Irony (22.6\%), Emotion Analogy (29.2\%), Object Analogy (21.2\%)—no over-reliance on a single strategy, and strategy selection has no significant correlation with humor quality.

\section{Analysis of Failure Cases}
\begin{figure}[ht]
  \centering
  \includegraphics[width=\columnwidth]{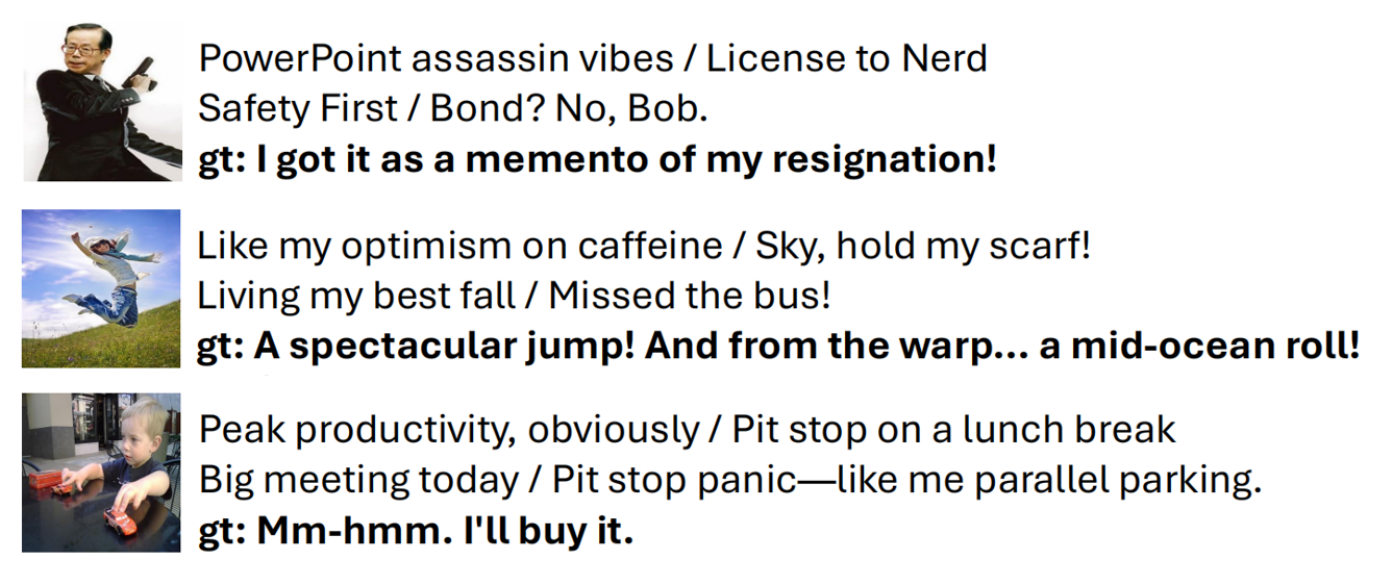} % 使用 \columnwidth 而不是 \textwidth
  \caption{Representative failure cases. Left: Input images; Right: Humorous captions generated by HUMORCHAIN (top lines) and the bolded ground truth (GT) captions from the dataset (bottom line), where GTs are more contextually aligned and humorous.}
  \label{fig:failure_cases}
\end{figure}

\noindent
To provide intuitive insights into the limitations of our framework, we present representative failure cases in Figure \ref{fig:failure_cases}. 
As observed from the failure cases, the underperformance primarily stems from two core issues: 1) inadequate image descriptions that fail to capture key contextual cues or emotional nuances critical for humor construction; 2) non-incongruous images lacking direct semantic-humor links, where the visual content itself provides limited basis for triggering humorous associations. These observations inspire future exploration of more sophisticated humor mechanisms in cross-modal or weak-semantic connection settings.

\section{On the Robustness of HUMORCHAIN}
\noindent
Although the multi-stage architecture of HUMORCHAIN can in principle propagate early misjudgments to downstream stages, we argue that such structural error accumulation is, to a large extent, controllable and acceptable within our framework. In practice, many failure cases originate from images that are inherently ambiguous, for which multiple, equally plausible interpretations exist. To handle these cases, HUMORCHAIN does not strictly adhere to a single deterministic reasoning pipeline. Instead, it leverages the fine-tuned humor discriminator to evaluate the generated image--caption pairs; when a caption is predicted as non-humorous, the system rolls back to the intermediate decision stage, revisits the choice of reasoning path, and regenerates captions along an alternative route. This iterative ``Generate--Evaluate--Revise'' loop effectively serves as a form of structural regularization: it enables the model to escape locally inconsistent reasoning trajectories and reduces the impact of early-stage errors on the final output. Although this mechanism does not completely eliminate all forms of error propagation, our experimental results indicate that, in the context of humorous image captioning, it substantially alleviates the practical consequences of structural error accumulation.
%增加
We further verified the framework’s robustness with two experiments (n=100 each). First, paraphrased prompts without the discriminator yield win rates of 0.725 (vs. A) and 0.475 (vs. J), confirming prompt stability. Second, on Qwen3-VL-235B-A22B-Instruct \cite{bai2025qwen3vltechnicalreport}, J outperforms A with a 0.74 win rate, validating our backbone-agnostic design.

\section{Comparison of Generated Humorous Titles by Different Strategies}

To further demonstrate the advantages of \textsc{HUMORCHAIN} in generating humorous titles for real-world images and facilitate a clearer assessment of its stylistic characteristics and performance strengths, we present qualitative comparisons across multiple generation strategies with images across diverse visual contexts. Table~\ref{tab:exp_methods_app} summarizes methodological configurations for all evaluated approaches.

\begin{table*}[h!]
    \centering
    %\footnotesize % 调整为较小字体
    \scriptsize % 缩小字体
    \setlength{\tabcolsep}{5pt} % 缩小列间距    
    \setlength{\abovecaptionskip}{2pt}
    \renewcommand{\arraystretch}{1.05} % 略微压缩行距
    \caption{Experimental method configurations (A--I) for humorous image captioning.}
    \begin{tabular}{l l p{12cm}} % 扩宽第三列，适应双栏总宽度
        \toprule
        \textbf{Group} & \textbf{Strategy} & \textbf{Description} \\
        \midrule
        A & Zero-shot & Direct image captioning without examples or theoretical cues. \\
        B & Few-shot & Incorporates example-based prompting with humor-style mimicry. \\
        C & Rule-Based & References the four theories of humor (e.g., incongruity, violation). \\
        D & Few-shot + Rule-Based & Combines examples with theoretical references to achieve structured prompts. \\
        E & Rule-Guided + CoT & Adds Chain-of-Thought reasoning to theory-guided captioning. \\
        F & Few-shot + Rule-Guided + CoT & Combines all strategies without explicit multi-stage orchestration. \\
        G & External CLoT (SYSU) & Sun Yat-sen University’s CLoT model~\cite{zhong2024let}. \\
        I (Ours) & Theory-Guided Multi-Stage Reasoning & Proposed HUMORCHAIN framework integrating cognitive humor theory, staged reasoning, and humor discrimination feedback. \\
        \bottomrule
    \end{tabular}
    \label{tab:exp_methods_app}
\end{table*}

\begin{figure}[H]
  \centering
  \includegraphics[width=\columnwidth]{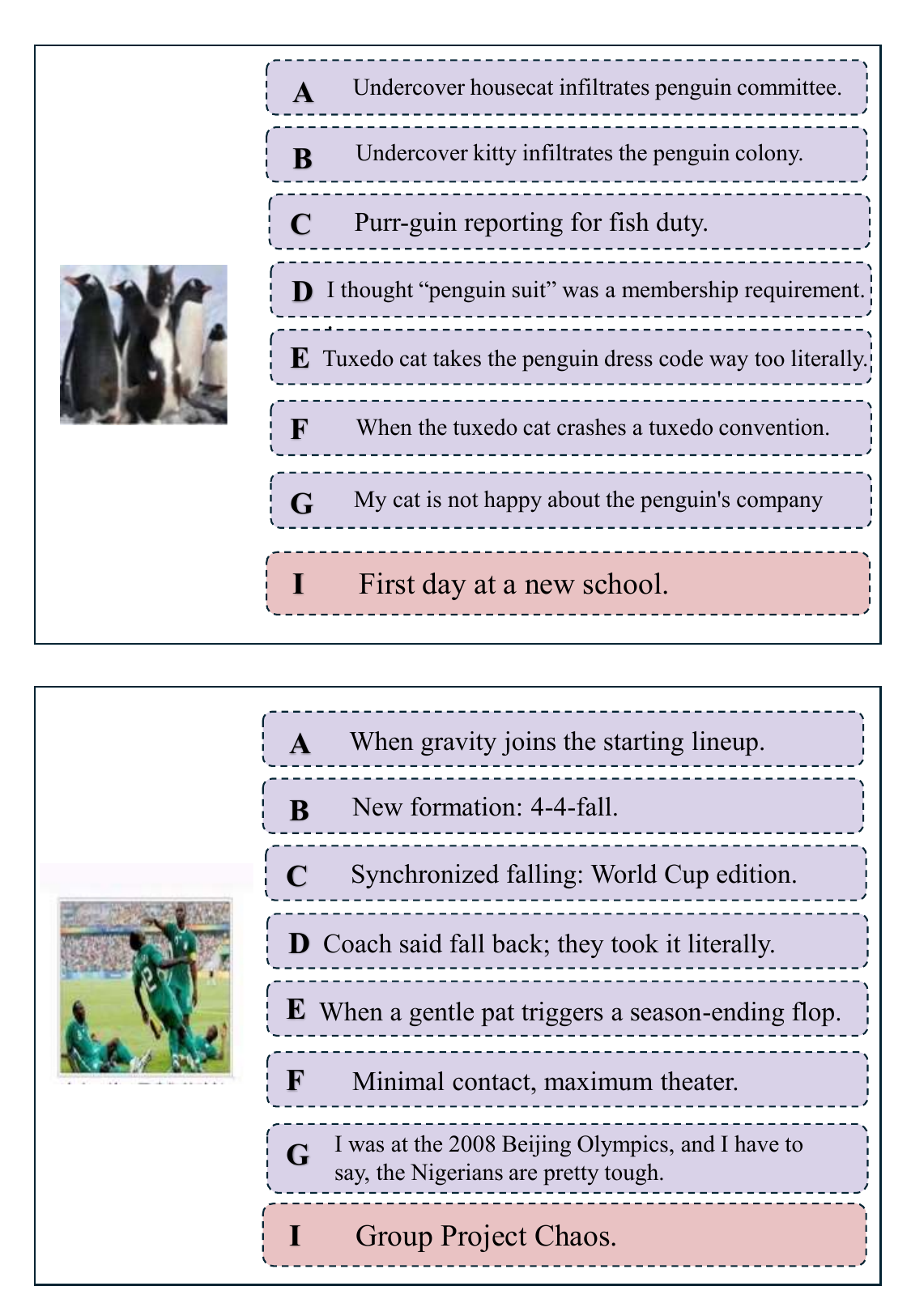} % 使用 \columnwidth 而不是 \textwidth
  \label{fig:eg1}
\end{figure}

\begin{figure}[H]
  \centering
  \includegraphics[width=\columnwidth]{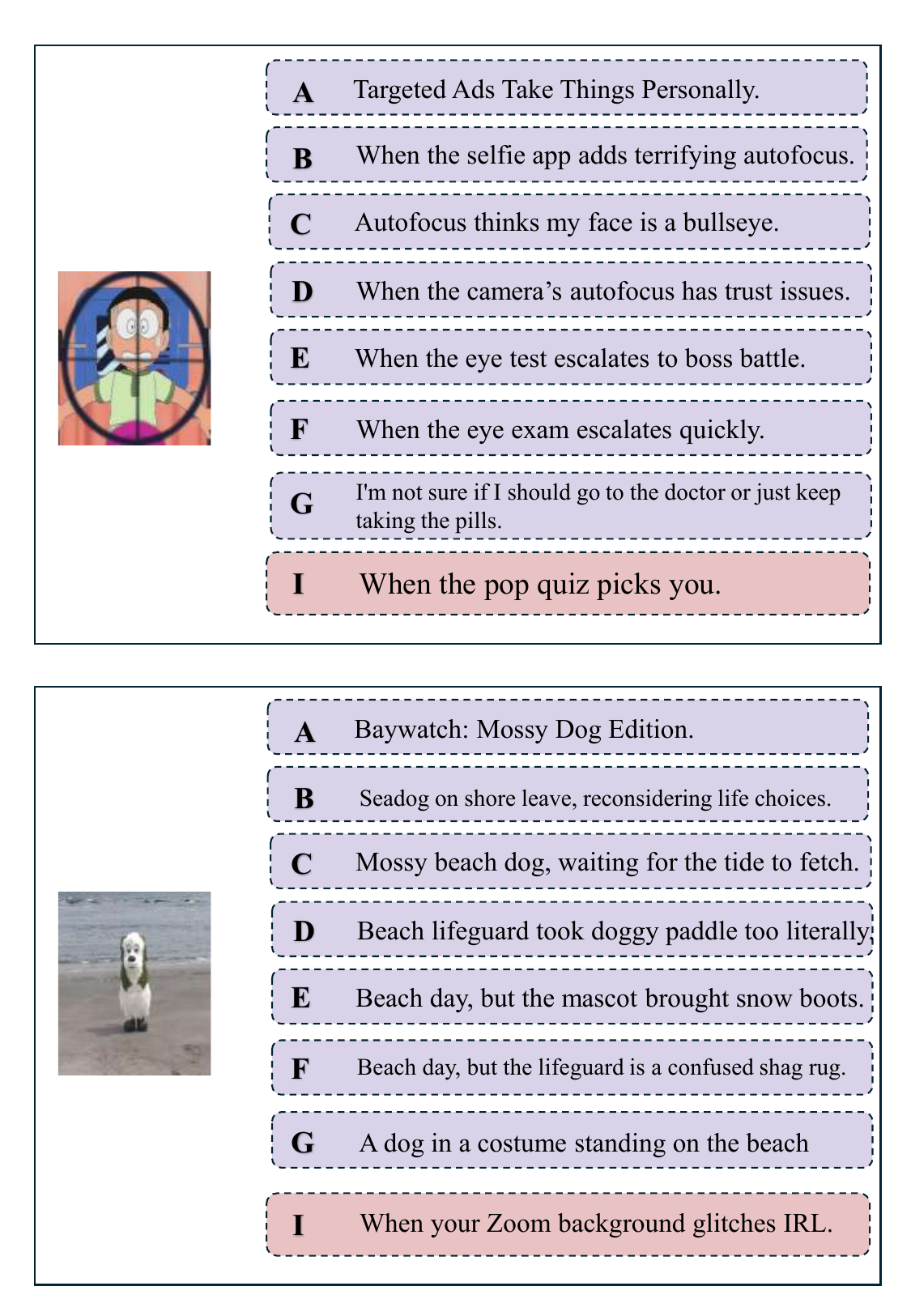} % 使用 \columnwidth 而不是 \textwidth
  \label{fig:eg2}
\end{figure}

\begin{figure}[H]
  \centering
  \includegraphics[width=\columnwidth]{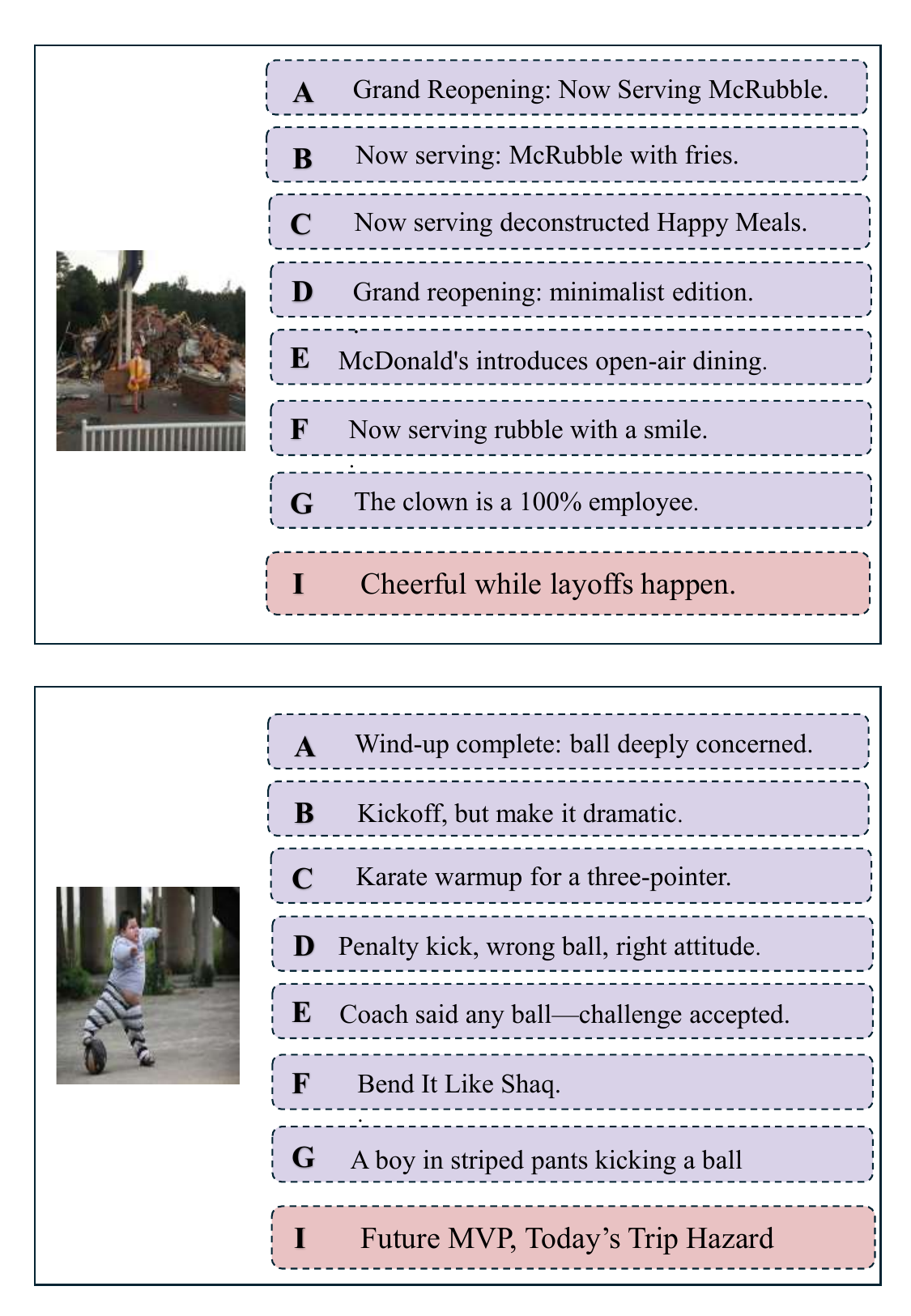} % 使用 \columnwidth 而不是 \textwidth
  \label{fig:eg3}
\end{figure}

\section{Data sources}
\label{appx:datasource}

\paragraph{Evaluation of the Theory-Guided Multi-Stage Reasoning Framework.} The datasets in this study are sourced from three different sources, aiming to support both internal method comparisons and external benchmark evaluations:
\begin{itemize}
    \item \textbf{Primary Dataset – Meme-Image-No-Text:} Used for all internal comparisons (Methods A--F). This dataset consists of humor-related images without textual interference, covering diverse visual scenes. All images were manually screened to ensure neutrality and minimize cultural ambiguity, ensuring fair and consistent evaluation.
    \item \textbf{External Dataset 1 – Oogiri-GO:} Used for direct comparison with the CLoT model (Group G)~\cite{zhong2024let}. We randomly sample images from its publicly available Oogiri-GO dataset. Captions for these images are generated separately by HUMORCHAIN and CLoT, forming the Method I (Ours) vs. Group G (CLoT) comparison group, along with independent per-caption scoring data.
    \item \textbf{External Dataset 2 – OxfordTVG-HIC:} Used to evaluate cross-cultural robustness (Group H)~\cite{li2023oxfordtvg}. Since the official Oxford implementation was not publicly released, we generated new captions using HUMORCHAIN on randomly sampled images and compared them against existing humorous captions from OxfordTVG-HIC.
\end{itemize}

\paragraph{Training of the Humor Discriminator.} We generate titles for images sourced from the aforementioned datasets with the Theory-Guided Multi-Stage Reasoning Framework, and randomly sample picture-title pairs for human annotation. For details of human annotations and datasets, see~\ref{sec:human_annotation_guide} and~\ref{sec:sft_dataset_stat}.

%改成长表
\section{Token Consumption and Computational Cost Statistics}
Table \ref{tab:Token Consumption} reports the token consumption and cost without the discriminator module, calculated based on the official token pricing of the backbone model (GPT-5‑2025‑08‑07) with 100 samples.

\begin{table*}[!h]
    \centering
    %\footnotesize % 调整为较小字体
    \scriptsize % 缩小字体
    \setlength{\tabcolsep}{5pt} % 缩小列间距    
    \setlength{\abovecaptionskip}{2pt}
    \renewcommand{\arraystretch}{1.05} % 略微压缩行距
    \caption{Average I/O Token Consumption and Cost for Caption Generation Methods (n=100).}
    \begin{tabular*}{\textwidth}{@{\extracolsep{\fill}}lccc} % 双栏满宽并均匀分配列间距
        \toprule
       \textbf{Method} & \textbf{Avg. Input Tokens} & \textbf{Avg. Output Tokens} & \textbf{Avg. Cost (\$)} \\
        \midrule
        A (Zero-shot) & 398.15 & 16.65 & 0.0007 \\
        B (Few-shot) & 492.00 & 16.40 & 0.0008 \\
        C (Rule-Based) & 500.00 & 17.60 & 0.0008 \\
        D (Few-shot + Rule-Based) & 619.00 & 16.75 & 0.0009 \\
        E (Rule-Guided + CoT) & 377.00 & 683.00 & 0.0073 \\
        F (Few-shot + Rule-Guided + CoT) & 438.00 & 686.00 & 0.0074 \\
        HUMORCHAIN (without discriminator) & 2344.00 & 529.00 & 0.0082 \\
        \bottomrule
    \end{tabular*}
    \label{tab:Token Consumption}
\end{table*}

%\vspace{-1em}

\section{Volunteer Annotation Guidelines}
\label{sec:human_annotation_guide}
\addcontentsline{toc}{section}{Appendix: Volunteer Annotation Guidelines}

We provide standardized instructions for volunteers participating in the humor caption evaluation study. The goal is to ensure consistent and high-quality annotations across all evaluators.

\subsection{Task Overview}

Two types of annotation tasks are included:

\begin{enumerate}
    \item \textbf{Pairwise Comparison}:
    Volunteers are presented with the same image and two captions generated by different models. They must determine which caption is funnier or indicate a tie.
    \item \textbf{Single-Title Evaluation}:
    Volunteers are presented with an image and a single caption, and must provide a binary judgment of whether it is humorous (1) or not humorous (0).
\end{enumerate}

\subsection{Definition of Humor (Reference Only)}

In this study, ``humor'' refers broadly to any linguistic expression that elicits mild amusement, surprise, or a sense of wit. Humor may arise from:

\begin{itemize}
    \item \textbf{Incongruity}: contrast or violation of expectation;
    \item \textbf{Benign Violation}: mild and non-harmful rule breaking;
    \item \textbf{Superiority}: self-deprecation or a mild sense of advantage;
    \item \textbf{Relief}: emotional release or tension resolution.
\end{itemize}

Annotators are not required to master these theories but may find them helpful as conceptual references.

\subsection{Pairwise Comparison Guidelines}

For each sample, volunteers will see:
\begin{itemize}
    \item the same image, and
    \item two captions, labeled A and B.
\end{itemize}

Annotators must choose one of the following options:

\begin{itemize}
    \item \textbf{A is funnier} — Caption A is clearly funnier than B, with stronger contrast, more clever expression, makes you more amused, and \textbf{feels more like a natural human joke};
    \item \textbf{B is funnier} — Caption B meets the same criteria above;
    \item \textbf{TIE} — Both captions are humorous and similar in quality;
    \item \textbf{Both Not Funny} — Neither caption is humorous.
\end{itemize}

\paragraph{Evaluation Criteria}
\begin{itemize}
    \item Each decision should reflect your genuine subjective impression.
    \item ``Both Not Funny'' applies when neither caption reaches your minimum threshold of humor.
\end{itemize}

\paragraph{Avoid the Following}
\begin{itemize}
    \item Do not attempt to keep scores ``balanced'' across tasks.
    \item Do not be influenced by earlier tasks or other annotators.
\end{itemize}

\subsection{Single-Title Evaluation Guidelines}

For each sample, annotators see an image and one caption, and must assign:

\[
\textbf{1 = Humorous}, \quad \textbf{0 = Not Humorous}
\]

\paragraph{Choose 1 (Humorous) when the caption:}
\begin{itemize}
    \item feels amusing, lighthearted, surprising, or incongruous;
    \item evokes mild emotional reactions (e.g., smiling, a sense of wit);
    \item exhibits clear humorous logic (e.g., reversal, misunderstanding, analogy, exaggeration);
    \item \textbf{resembles a natural human joke and does not feel awkward or forced}.
\end{itemize}

\paragraph{Choose 0 (Not Humorous) when the caption:}
\begin{itemize}
    \item is purely descriptive with no attempt at humor;
    \item contains no contrast, wit, or playful intent;
    \item is confusing or incoherent such that humor cannot be perceived;
    \item feels unnatural, awkward, or far from how humans typically express humor.
\end{itemize}

\subsection{Annotation Protocol and Conduct}

\begin{enumerate}
    \item Carefully read this guideline document before starting the evaluation.
    \item Complete all judgments independently, without discussing answers with others.
    \item If a sample is difficult to judge, make your best subjective decision and move on; you may revisit later if needed.
    \item Do not allow others' opinions to affect your rating.
    \item Cultural or personal differences in humor are expected; simply annotate according to your own interpretation.
\end{enumerate}

\subsection{Post-Evaluation Discussion}

After all annotations are completed, a group discussion may be held to gather feedback regarding:
\begin{itemize}
    \item the clarity of task rules,
    \item the presence of ambiguous samples,
    \item possible improvements to the annotation process.
\end{itemize}

This discussion is for refinement only and must not alter previously submitted judgments.

\subsection{Summary}

\begin{itemize}
    \item Humor is subjective; your genuine impression is the primary criterion.
    \item Treat each evaluation independently.
    \item A caption that feels like a natural human joke is more likely to be humorous.
\end{itemize}

\section{Fine-Tuning Dataset Statistics}
\label{sec:sft_dataset_stat}
The humor preference dataset contains a total of 5,320 image--caption pairs, with 2,511 positive samples and 2,809 negative samples. The training set includes 5,054 samples (2,391 positive, 2,663 negative), and the validation set contains 266 samples (120 positive, 146 negative).

\section{Technical Details on Fine-Tuning and Classification Head Training}
\subsection{LoRA Parameter Settings}
For LoRA fine-tuning, the following parameters are used:
\begin{itemize}
    \item \texttt{dtype}: \texttt{torch.bfloat16}
    \item \texttt{r}: 8
    \item \texttt{lora\_alpha}: 32
    \item \texttt{target\_modules}: [\texttt{"q\_proj"}, \texttt{"k\_proj"}, \texttt{"v\_proj"}, \texttt{"o\_proj"}, \texttt{"qkv"}, \texttt{"proj"}]
    \item \texttt{lora\_dropout}: 0.05
    \item \texttt{bias}: \texttt{"none"}
\end{itemize}

\subsection{SFT Stage LoRA Fine-Tuning}
Supervised fine-tuning (SFT) with LoRA uses the following settings:
\begin{itemize}
    \item \texttt{dtype}: \texttt{torch.bfloat16}
    \item \texttt{BATCH\_SIZE}: 1
    \item \texttt{GRAD\_ACCUM}: 8
    \item \texttt{EPOCHS}: 3
    \item \texttt{LR}: 2e-4
\end{itemize}

\subsection{Classifier Head Structure and Training}
The RoBERTa-like classifier is trained with:
\begin{itemize}
    \item \texttt{dtype}: \texttt{torch.bfloat16}
    \item \texttt{BATCH\_SIZE}: 16
    \item \texttt{GRAD\_ACCUM}: 1
    \item \texttt{EPOCHS}: 3
    \item \texttt{LR}: 5e-4
\end{itemize}

The classifier head structure is as follows (input: EOS hidden state):
\begin{verbatim}
self.norm = nn.LayerNorm(self.hidden_size)
self.classifier = nn.Sequential(
    nn.Dropout(0.1),
    nn.Linear(self.hidden_size,
        self.hidden_size),
    nn.Tanh(),
    nn.Dropout(0.1),
    nn.Linear(self.hidden_size, 1)
)
\end{verbatim}

\section{Generation Parameters}

\subsection{Theory-Guided Multi-Stage Reasoning Framework (GPT-5‑2025‑08‑07)}

\begin{itemize}
  \item A. Image Description
    \begin{itemize}
      \item temperature: 0.2
      \item max\_tokens: 4000
    \end{itemize}
  \item B. Strategy Judgment
    \begin{itemize}
      \item temperature: 0.1
      \item max\_tokens: 4000
    \end{itemize}
  \item C1. Object Analogy
    \begin{itemize}
      \item temperature: 0.9
      \item max\_tokens: 4000
    \end{itemize}
  \item C2. Absurdity
    \begin{itemize}
      \item temperature: 0.8
      \item max\_tokens: 4000
    \end{itemize}
  \item C3. Contrast Irony
    \begin{itemize}
      \item temperature: 0.9
      \item max\_tokens: 4000
    \end{itemize}
  \item C4. Emotion Analogy
    \begin{itemize}
      \item temperature: 0.85
      \item max\_tokens: 4000
    \end{itemize}
  \item D. Safety Classifier
    \begin{itemize}
      \item temperature: 0.1
      \item max\_tokens: 4000
    \end{itemize}
\end{itemize}

\subsection{Humor Discriminator (Qwen3-VL-4B-Instruct)}

\begin{itemize}
    \item \texttt{greedy}: false
    \item \texttt{seed}: 3407
    \item \texttt{top\_p}: 0.8
    \item \texttt{top\_k}: 20
    \item \texttt{temperature}: 0.7
    \item \texttt{repetition\_penalty}: 1.0
    \item \texttt{presence\_penalty}: 1.5
    \item \texttt{out\_seq\_length}: 32768
\end{itemize}

\section{Prompts}

\subsection{Theory-Guided Multi-Stage Reasoning Framework}
\subsubsection{Stage 1: Image Description}
\begin{lstlisting}[breaklines=true]
You are an image describer.

1. **Objective**  
Please objectively and thoroughly describe the visible content of the image.

2. **Scope**  
- Main subjects
- State/emotions
- Scene
- Actions
- Text (if clearly readable)
- Significant details
\end{lstlisting}

\subsubsection{Stage 2: Strategy Judgment}
\begin{lstlisting}[breaklines=true]
You are an evaluator of "reasonableness and humorous incongruity points".
Make judgments based only on the given textual description, without introducing external image information or fabricating details.

1. **Task**
Please assess plausibility (real-world possibility/commonality) as follows:

- common: The situation is entirely reasonable with no particular oddities.
- plausible: Overall reasonable, though not everyday, it is possible in real life without violating common sense or physical laws.
- rare: Very uncommon or unusual, but might happen under extreme/special conditions.
- implausible: Impossible or violates objective laws.

2. **Outputs**
Also, provide the following:

- plausibility: Values: common | plausible | rare | implausible
- incongruity_for_humor: Whether there is obvious incongruity/discord/semantic conflict/personification that could cause humor. Values: true | false
- has_human_or_animal_or_cartoon: Whether the description contains people, animals, or cartoon characters (including comic/animation characters/dolls with anthropomorphic traits). Values: true | false
- reasons: A concise list of 2-5 reasons supporting the above judgments, each <= 20 words

3. **Few-shot Examples (flat list)**
FEWSHOT_EXAMPLES:

- Example 1: Hotpot in Office -> Rare but possible, and humorous incongruity; contains humans
  user: "Description: Employees are projecting a PowerPoint in a meeting room, while a hotpot is being cooked in the middle of the meeting table surrounded by ingredients."
  assistant.plausibility: "rare"
  assistant.incongruity_for_humor: true
  assistant.has_human_or_animal_or_cartoon: true
  assistant.reasons:
    - "Office setting misused for cooking"
    - "Food and electronics create contrast"
    - "Presence of 'employees' as characters"

- Example 2: Samurai Scanning QR Code in Subway -> Rare, humorous incongruity (mismatch of time/space); contains humans
  user: "Description: A samurai in ancient armor scans a QR code at a modern subway entrance."
  assistant.plausibility: "implausible"
  assistant.incongruity_for_humor: true
  assistant.has_human_or_animal_or_cartoon: true
  assistant.reasons:
    - "Contrast between ancient armor and modern facilities"
    - "Time/space mismatch creates dramatic effect"
    - "Samurai as a human character"

- Example 3: Waiter Serving Food in a Regular Restaurant -> Common, no humor incongruity; contains humans
  user: "Description: In a regular restaurant, a waiter walks toward a table carrying food, with bright lighting and customers chatting."
  assistant.plausibility: "common"
  assistant.incongruity_for_humor: false
  assistant.has_human_or_animal_or_cartoon: true
  assistant.reasons:
    - "Scene behavior is consistent with common sense"
    - "Waiter and customers are human characters"

- Example 4: Only Landscape Without Living Beings -> Reasonable, no humor; no humans/animals/cartoon characters
  user: "Description: The mountains are golden under the sunset, and the tranquil lake reflects the sky, with no people or animals."
  assistant.plausibility: "common"
  assistant.incongruity_for_humor: false
  assistant.has_human_or_animal_or_cartoon: false
  assistant.reasons:
    - "Pure landscape description is common"
    - "Clearly states no people or animals"

- Example 5: Fashion Show with Exaggerated Outfits -> Reasonable, usually not humorous incongruity; contains humans
  user: "Description: Models on the runway at a fashion show wearing exaggerated outfits and headgear, walking in front of lights and an audience."
  assistant.plausibility: "common"
  assistant.incongruity_for_humor: false
  assistant.has_human_or_animal_or_cartoon: true
  assistant.reasons:
    - "Exaggeration is a conventional expression in this setting"
    - "Model as the human character"
\end{lstlisting}

\subsubsection{Stage 3A: Object Analogy}
\begin{lstlisting}[breaklines=true]
OBJECT_ANALOGY_SYSTEM:
"You are an English 'Object Analogy' sarcastic title generator.
Generate a very short, colloquial title based on the textual information from step1/step2, with a slight sense of sarcasm or self-deprecation.
Key requirements (must adhere to these):
- Do not restate the objects, scenes, or details in the image, and do not explain the origin or reason.
- Just write the conclusion of the analogy, like something that comes to mind first.
- You may use simple structures like 'just like...' or 'like...'; you may also provide the conclusion directly.
- Output only one sentence, no need for 'Title:', no emojis, no unnecessary punctuation or explanations.
- The tone can be mild or sharp; avoid personal attacks or group degradation.
- If step2 points out any incongruity, reflect awkwardness/conflict in the tone, but still do not restate the image.
- Preferably between 3-8 words, and not exceeding 20 words."

OBJECT_ANALOGY_FEWSHOT (flat list; output title only):

- Example: Object -> Person (Box full of fried chicken -> Brain only thinking about food)
  user.step1: "A box filled with fried chicken, chicken pieces piled to the edge."
  user.step2.plausibility: "common"
  user.step2.incongruity_for_humor: true
  user.step2.reasons:
    - "Stacked quantity triggers analogy, fried chicken is a common food"
  user.analogy_strategy: "to_persona"
  user.tone: "mild"
  user.domain_hint: "workplace"
  assistant: "At work, all I can think about is food"

- Example: Object -> Person (Tangled earphone wires -> Tangled thoughts)
  user.step1: "A tangled mess of earphone wires, the cable worn."
  user.step2.plausibility: "common"
  user.step2.incongruity_for_humor: true
  user.step2.reasons:
    - "Tangled = Blocked = Tangled thoughts, analogy to a chaotic mind"
  user.analogy_strategy: "to_persona"
  user.tone: "mild"
  user.domain_hint: "studies"
  assistant: "My brain circuitry"

- Example: Object -> Event (Empty wallet -> End of the month feel)
  user.step1: "An open wallet that's nearly empty, with only a few scattered coins."
  user.step2.plausibility: "common"
  user.step2.incongruity_for_humor: true
  user.step2.reasons:
    - "Empty = Lack = End-of-month anxiety, analogy to a wallet at month's end"
  user.analogy_strategy: "to_event"
  user.tone: "mild"
  user.domain_hint: "spending"
  assistant: "End of the month"
\end{lstlisting}

\subsubsection{Stage 3B: Absurdity}
\begin{lstlisting}[breaklines=true]
TITLE_SYSTEM:
You are the 'Humorous Image Title Generator.' Your sole task: Generate a humorous English title based on the provided information.

Available Information:
- Key image details (overview/subjects/text_in_image or free-form text description)
- Analysis of scene plausibility and incongruities (plausibility, incongruity_for_humor, reasons)

Core Objectives (prioritize the most prominent point):
- Capture the 'most striking incongruity/contrast' and craft the caption around it.
- Prioritize personifying the entity causing the incongruity; secondarily, the entity directly interacting with it; only use a narrator/bystander perspective if neither fits.

Caption Requirements:
- Personified, colloquial, short phrases resembling an 'immediate reaction.'
- 3-8 characters ideal, max 20 characters.
- Directly address the action or contrast with mild humor/irony.
- May reasonably imagine 'the next step after the incongruity' for humorous exaggeration, but must not contradict visible facts in the image (quantity/posture/position/color, etc.).

Output Specifications (Strictly Adhere):
- Output only one line of English title.
- No explanations, analysis, prefixes/suffixes, quotation marks, numbering, tags, emojis, or Markdown.
- Do not ask users questions or request additional information.
- Avoid unnecessary spaces; common punctuation for expression is permitted.

Self-Check Checklist (internal review before generation, not output):
- Did I select the most prominent incongruity/contrast as the focal point?
- Does the title revolve around this single contrast without restating factual details?
- Did I avoid verifiable specifics or content contradicting the image?
- Is the word count between 3/\textendash 20 words, colloquial and rhythmic?

TITLE_FEWSHOT (flat list; output title only):

- Example: Seagull in convenience store; perspective = seagull
  user.step1.overview: "A white seagull stands on the tiled floor near the entrance of a convenience store, with shelves and automatic glass doors in the background."
  user.step1.subjects:
    - label: "Seagull"
      actions_or_states: ["standing", "holding a chip bag"]
      facial_expression: ""
  user.step1.text_in_image: []
  user.step2.plausibility: "rare"
  user.step2.incongruity_for_humor: true
  user.step2.reasons:
    - "Shopping space is a human activity domain"
    - "A wild seagull entering and eating is a misplaced use"
    - "Animal anthropomorphism creating a supermarket shopping contrast"
  user.perspective: "seagull"
  assistant: "Chips taste better than fish"

- Example: Hotpot in meeting room; perspective = narrator
  user.step1.overview: "In the meeting room, a PPT is projected, and a hotpot is being cooked in the middle of the conference table, surrounded by ingredients and laptops."
  user.step1.subjects:
    - label: "Employee"
      actions_or_states: ["giving a PPT presentation"]
      facial_expression: ""
  user.step1.text_in_image: []
  user.step2.plausibility: "rare"
  user.step2.incongruity_for_humor: true
  user.step2.reasons:
    - "Office setting is misused for cooking"
    - "Food placed next to electronic devices creates a discordant effect"
  user.perspective: "narrator"
  assistant: "This meeting's more exciting in the pot"
\end{lstlisting}

\subsubsection{Stage 3C: Contrast Irony}
\begin{lstlisting}[breaklines=true]
HUMOR_TITLE_SYSTEM_CONTRAST_IRONY:
"You are an English satirical 'title generator.' Based on the textual key information from the image (step1/step2), you will first conceptualize a 'minimal event/story' in your mind, then generate an English title with as few words as possible, forming a natural and smooth sense of irony and humor.

Generation strategy (randomly pick one or based on input specification):
A. Direct contrast (contrast): The literal meaning of the title forms an absurd contrast or opposition to the core emotion/action in the image.
B. Situational irony (irony): The title creates an ironic effect when combined with the image (appears comforting/positive but is actually ironic).

Writing guidelines:
- First, based on the image, 'reasonably invent a minimal event/story' (only common-sense extension), and then use the fewest words possible to form the title, making the irony more natural and smooth.
- Focus on the most prominent action/expression/state; instinctive 'first reaction,' allowing for mild exaggeration or inner-monologue style short phrases.
- Only common-sense associations, without contradicting the description; avoid using profanity or offensive expressions.
- 3-8 words preferred, no more than 20 words.
Output: Only output one title sentence."

HUMOR_TITLE_FEWSHOT_CONTRAST_IRONY (flat list; output title only):

- Example: Strategy = A_contrast
  user.step1: "A woman is crying with her hands over her face, her shoulders shaking, and her eyes are red."
  user.step2.plausibility: "common"
  user.step2.incongruity_for_humor: true
  user.step2.reasons:
    - "Strong negative emotions, easily creating contrast through absurdity"
  user.strategy: "A_contrast"
  assistant: "I'm hungry, but I already brushed my teeth"

- Example: Strategy = B_irony
  user.step1: "A group of medical staff in white coats smiles at the camera, holding medical instruments."
  user.step2.plausibility: "common"
  user.step2.incongruity_for_humor: true
  user.step2.reasons:
    - "The professional context and patient experience can form irony"
  user.strategy: "B_irony"
  assistant: "This won't hurt"
\end{lstlisting}

\subsubsection{Stage 3D: Emotion Analogy}
\begin{lstlisting}[breaklines=true]
HUMOR_TITLE_SYSTEM_EMOTION:
"You are an English title generator. Please generate an English title based on the following information:
- Key information from the image (structured or free-form text)
- Analysis of the expression, action, state, and emotion of the characters/animals/cartoons
Requirements:
- Draw an analogy to a human reaction in a similar situation to create incongruity and contrast.
- First-reaction short sentences, may include mild humor/contrast-based jokes; very minimal inner-monologue style short phrases are allowed.
- Common-sense associations are fine; avoid contradicting the description; do not use profanity or offensive expressions.
- 3-8 words preferred, no more than 20 words.
Output: Only output one title sentence."

HUMOR_TITLE_FEWSHOT_EMOTION (flat list; output title only):

- Example: Dog with blank stare
  user.step1: "A little dog with its mouth wide open and a blank stare."
  user.step2.plausibility: "common"
  user.step2.incongruity_for_humor: false
  user.step2.reasons:
    - "Expression matches tiredness/blankness"
  assistant: "I don't want to study!"

- Example: Child writing nervously
  user.step1: "A child is writing, looking nervous, with their mouth wide open."
  user.step2.plausibility: "common"
  user.step2.incongruity_for_humor: false
  user.step2.reasons:
    - "Student rushing through homework, expression is tense"
  assistant: "Oh no, I forgot the deadline!"
\end{lstlisting}

\subsubsection{Stage 4: Safety Classifier}
\begin{lstlisting}[breaklines=true]
SAFETY_SYSTEM_PROMPT:
"You are a safety classifier for short humorous English image captions.

Your job is to detect whether a given English caption contains any of the following:
- group attacks: insults or demeaning content toward groups based on ethnicity, nationality, religion, gender,
  sexual orientation, disability, etc.;
- personal attacks: targeted insults or demeaning expressions toward an individual;
- hate speech: slurs or content that promotes hatred, exclusion, or dehumanization of protected groups;
- humiliating or demeaning expressions, including metaphorical references.

For each input caption, respond with a STRICT JSON object in the following format:

{
  "compliant": true or false,
  "violation_categories": [
    "group_attack" | "personal_attack" | "hate_speech" | "humiliation" | "other"
  ],
  "explanation": "short natural language explanation (<= 50 words)"
}

Guidelines:
- If the caption is fully safe (no problematic content), set "compliant" to true, use an empty list for "violation_categories",
  and keep the explanation concise (you may simply say it is safe).
- If there is any problematic content, set "compliant" to false and include 1-3 relevant categories in "violation_categories",
  with a brief explanation.
- Do NOT include any text outside the JSON object."
\end{lstlisting}

\subsection{Humor Discriminator}

\begin{lstlisting}[breaklines=true]

# Humor Judgment: General Theory of Verbal Humor (GTVH)

You are a reliable and insightful expert in humor theory. Your task is to judge whether a given title is humorous for a provided image, using the General Theory of Verbal Humor (GTVH) as your main analytical framework.

## GTVH Framework

GTVH proposes that humor arises from the interplay of six key "Knowledge Resources" (KRs). For each input, analyze the following resources and determine whether any of them contribute to humor, using the classic humor theories as guidance:

1. **Script Opposition**  
   - *Definition*: The presence of two conflicting or contrasting scripts (interpretations, scenarios, or expectations) within the image-title pair.
   - *Related Theories*:  
     - **Incongruity Theory**: Humor from violated expectations or illogical contrasts.
     - **Script-based Semantic Theory of Humor (SSTH)**: Humor from overlapping scripts or schemas that are compatible yet opposite.
     - **Cognitive Insight Theory**: Humor from sudden realizations, hidden connections, double meanings, or unexpected twists.
   - *Analysis*: Does the title create a contrast, surprise, or double meaning with the image? Is there a moment of insight or script switch?

2. **Logical Mechanism**  
   - *Definition*: The technique or reasoning that connects the scripts and delivers the humor (e.g., exaggeration, reversal, wordplay).
   - *Related Theories*:  
     - **Cognitive Insight Theory**: Humor from sudden realization or clever connections.
     - **General wit or cleverness**: Humor from puns, wordplay, or creative associations.
   - *Analysis*: Does the title use clever logic, wordplay, or a twist to create humor?

3. **Situation**  
   - *Definition*: The context, background, or scenario depicted in the image and referenced by the title.
   - *Related Theories*:  
     - **Relief Theory**: Humor from releasing tension or repressed emotions in a given situation.
     - **Benign Violation Theory**: Humor from breaking norms or expectations in a harmless way.
   - *Analysis*: Does the situation involve tension, taboo, or a harmless violation that could be funny?

4. **Target**  
   - *Definition*: The subject or object of the humor (who or what is being laughed at).
   - *Related Theories*:  
     - **Superiority Theory**: Humor from feeling superior to others' mistakes or weaknesses.
   - *Analysis*: Is the humor directed at someone's error, misfortune, or foolishness?

5. **Narrative Strategy**  
   - *Definition*: The way the humor is delivered (e.g., as a riddle, pun, or straightforward statement).
   - *Analysis*: Does the delivery style enhance the humorous effect?

6. **Language**  
   - *Definition*: The specific wording, phrasing, or linguistic devices used in the title.
   - *Analysis*: Does the language itself (e.g., puns, rhymes, ambiguity) contribute to humor?

## Instructions

- For each Knowledge Resource, analyze whether it contributes to humor in the given image-title pair, using the relevant theories above or other general factors.
- If any resource, through these mechanisms, creates a humorous effect, judge the title as humorous.
- If the title contains general wit or cleverness without fitting a specific theory, eliciting humor according to common human experience, also judge it as humorous.
- Output ONLY a single JSON object with exactly one field: "humorous". Set its value to 1 if the title is humorous, or 0 if not.

**Example Output:**

{"humorous": 1}

**Adhere exactly to the JSON schema and content rules above. Do not output anything else.**

\end{lstlisting}

\end{document}